\pdfoutput=1
\documentclass[11pt]{article}

\usepackage{acl}

\usepackage{times}
\usepackage{latexsym}

\usepackage[T1]{fontenc}

\usepackage[utf8]{inputenc}

\usepackage{microtype}

\newcommand{\extodo}[1]{{\color{red!70!black}\textbf{[TODO: #1]}}}

\usepackage{adjustbox}

\usepackage{amsmath,amsfonts,bm}
\usepackage{amsthm}

\usepackage{makecell}

\usepackage{amssymb,mathtools}
\usepackage{bbm}
\usepackage{nccmath}

\DeclareMathOperator{\btheta}{\bm{\theta}}
\DeclareMathOperator{\bpr}{\mathbf{pr}}
\DeclareMathOperator{\bd}{\mathbf{d}}
\DeclareMathOperator{\bv}{\mathbf{v}}

\usepackage{tikz}
\usepackage{tikz-dependency}
\newcommand*\circled[1]{\tikz[baseline=(char.base)]{
            \node[shape=circle,draw,inner sep=.6pt] (char) {#1};}}

\usepackage{fontawesome5}
\usepackage{graphicx}
\usepackage{booktabs}

\usepackage{tcolorbox}
\tcbuselibrary{breakable,xparse,skins}
\usepackage{listings}
\usepackage{subcaption}
\usepackage{arydshln}

\definecolor{codegreen}{HTML}{006600}
\definecolor{codegray}{rgb}{0.5,0.5,0.5}
\definecolor{codepurple}{rgb}{0.58,0,0.82}
\definecolor{backcolour}{HTML}{f0f5f5}
\definecolor{agentcolor}{HTML}{993399}
\definecolor{clientcolor}{HTML}{333333}
\definecolor{referencecolor}{HTML}{1174ad}
\definecolor{rougecolor}{HTML}{B85450}
\definecolor{textbgcolor}{HTML}{F5F5F5}
\definecolor{promptbgcolor}{HTML}{ecf9ec}
\definecolor{workflowbgcolor}{HTML}{e6ecff}
\definecolor{slotcolor}{HTML}{cc7a00}
\definecolor{specialtokencolor}{HTML}{b30000}

\lstdefinestyle{mystyle}{
    backgroundcolor=\color{backcolour},   
    commentstyle=\color{codegreen},
    keywordstyle=\color{magenta},
    numberstyle=\tiny\color{codegray},
    stringstyle=\color{codepurple},
    basicstyle=\ttfamily\footnotesize,
    breakatwhitespace=false,         
    breaklines=true,                 
    captionpos=b,                    
    keepspaces=true,                 
    numbersep=5pt,                  
    showspaces=false,                
    showstringspaces=false,
    showtabs=false,                  
    tabsize=2
}

\newcommand{\agent}[2]{\begin{flushleft} \textcolor{agentcolor}{#1} \raisebox{-0.05cm}{\includegraphics[width=0.4cm]{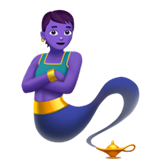}}\textcolor{agentcolor}{:}\ \textcolor{agentcolor}{#2}\end{flushleft} }
\newcommand{\client}[2]{\begin{flushright} \textcolor{clientcolor}{#1} \raisebox{-0.05cm}{\includegraphics[width=0.4cm]{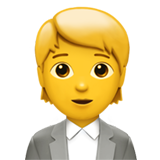}}\textcolor{clientcolor}{:} \ \textcolor{clientcolor}{#2}\end{flushright}}
\newcommand{\thumbsup}{\raisebox{-0.05cm}{\includegraphics[width=0.4cm]{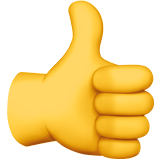}}}
\newcommand{\thumbsdown}{\raisebox{-0.05cm}{\includegraphics[width=0.4cm]{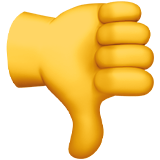}}}
\newcommand{\questionmark}{\raisebox{-0.05cm}{\includegraphics[width=0.4cm]{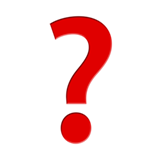}}}

\usepackage{tcolorbox}

\usepackage{xurl}
\usepackage[noabbrev,capitalize,nameinlink]{cleveref}

\title{Bootstrapping LLM-based Task-Oriented Dialogue Agents via Self-Talk}

\author{Dennis Ulmer\textsuperscript{\thanks{\hspace{0.2cm}Work conducted during an internship at Amazon.}\hspace{0.1cm}, \faCompass, \faFlag} \hspace{.5em}  Elman Mansimov\textsuperscript{\faAmazon} \hspace{.5em} \textbf{Kaixiang Lin}\textsuperscript{\faAmazon} \hspace{.5em} \textbf{Justin Sun}\textsuperscript{\faAmazon}\\ \textbf{Xibin Gao}\textsuperscript{\faAmazon} \hspace{.5em} \textbf{Yi Zhang}\textsuperscript{\faAmazon}\\
       \textsuperscript{\faCompass} IT University of Copenhagen
       \textsuperscript{\faFlag} Pioneer Centre for Artificial Intelligence\\
       \textsuperscript{\faAmazon} AWS AI Labs\\
        \texttt{mansimov@amazon.com}}

\begin{document}
\maketitle

\begin{abstract}
    Large language models (LLMs) are powerful dialogue agents, but specializing them towards fulfilling a specific function can be challenging. 
    Instructing tuning, i.e. tuning models on instruction and sample responses generated by humans \citep{ouyang2022training}, has proven as an effective method to do so, yet requires a number of data samples that a) might not be available or b) costly to generate.
    Furthermore, this cost increases when the goal is to make the LLM follow a specific workflow within a dialogue instead of single instructions. 
    Inspired by the self-play technique in reinforcement learning and the use of LLMs to simulate human agents, we propose a more effective method for data collection through LLMs engaging in a conversation in various roles. This approach generates a training data via "self-talk" of LLMs that can be refined and utilized for supervised fine-tuning. 
    We introduce an automated way to measure the (partial) success of a dialogue. This metric is used to filter the generated conversational data that is fed back in LLM for training. Based on our automated and human evaluations of conversation quality, we demonstrate that such self-talk data improves results.
    In addition, we examine the various characteristics that showcase the quality of generated dialogues and how they can be connected to their potential utility as training data.
    
\end{abstract}

\section{Introduction}

LLMs have recently proven their viability to be deployed as conversational agents in a variety of scenarios, promising novel possibilities in user-facing applications, such as working as a personal assistant or customer support. 
In these scenarios, dialogue agents have to be able to quickly adapt to new tasks, which in the current paradigm implies either the use of already-existing or the creation of new datasets. 
Naturally, this can become a prohibitive constraint in terms of feasibility, time, and resources. 
Furthermore, avoiding finetuning by using prompting to elicit a certain behavior can prove brittle and unreliable - while it still might work in some instances, there is no guarantee for robustness or consistency.\\

\begin{figure}[tb!]
    \centering
    \includegraphics[width=0.85\columnwidth]{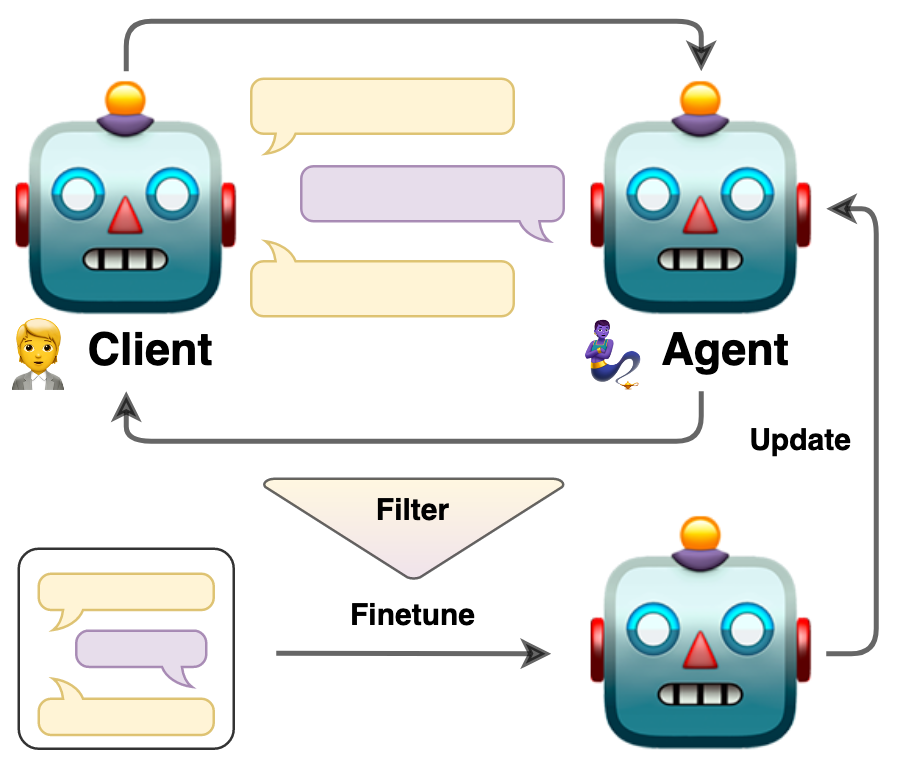}
    \caption{Schematic representation of our approach. Two LLMs, called a \emph{client} and an \emph{agent}, are prompted to converse with each other in different roles, with the agent asked to follow a specific narrative structure. Generated conversations will then be filtered by quality and used for supervised finetuning on the agent model until it adapts to the intended dialogue structure.}\label{fig:schema}
\end{figure}

In reinforcement learning (RL) for games, self-play has proven to be an effective mean to bootstrap powerful systems that can beat human experts \citep{silver2016mastering}, even when no initial human-generated data \citep{silver2017mastering} or rules \citep{schrittwieser2020mastering} are supplied. 
At the same time, increasingly sized language models trained with growing datasets have shown to develop limited but notable extents of language abilities \citep{hu2023prompt} and some degree of \emph{theory-of-mind}, i.e. reasoning over other agents, their behavior and the world state \citep{sap2022neural}. 
This has prompted studies in which they simulate single users \citep{liu2022generative, argyle2023out} or even interacting groups of agents \citep{liu2023training, park2023generative}. 
Recent work by \citet{gulcehre2023reinforced} also demonstrated how LLMs can be iteratively improved on their own, generating data for language modeling.
We take these developments as encouragement to explore LLM self-\emph{talk} as a way to bootstrap task-specific finetuning data from LLMs themselves. 
Specifically, we focus on task-oriented dialogue, a very relevant problem for applications such as customer service, pair programming and digital personal assistants.
We propose the following procedure: 
Two versions of an LLMs take part in a dialogue, impersonating a client and an agent. 
Both are given a description of how to act within the dialogue: 
The client is given a persona and motivation, while the agent is equipped with a fixed set of behaviors, depending on the request of the agent. 
They then engage in a conversation, which is recorded as a potential training sample. 
Some of these conversations will be successful, however, many of them will not be useful, since they will be off-topic or of low-quality. 
For this reason, we employ an additional filtering step to avoid training the model on counterproductive samples. 
A schematic depiction of our method is shown in \cref{fig:schema}.
Our contributions are as follows: \circled{1} We propose a new method to bootstrap training data for task-oriented dialogue agents. \circled{2} We explore different quality assurance options. \circled{3} We also introduce automated evaluation metrics to evaluate dialogue success and conversational consistency and \circled{4} assess our trained agents in a human evaluation study.

\section{Related Work}

\paragraph{Self-Play} has been used extensively in RL for games to bootstrap powerful agents: 
Beginning with early successes in playing backgammon \citep{tesauro1994td, pollack1996did} and later Othello \citep{van2013reinforcement}, further breakthroughs were not achieved until the arrival of AlphaGo \citep{silver2016mastering}, which was pre-trained on human expert moves before self-play. 
Following iterations of the system forwent the pre-training \citep{silver2017mastering} and later even achieved super-human abilities on multiple games even without explicit knowledge of the rules \citep{schrittwieser2020mastering}. 
In NLP, there has been some work with self-play for dialogue systems \citep{shah2018bootstrapping, shah2018building}, but these date back before the advent of conversational LLMs. 
Instead, these approaches rely on pre-defined task schemata and slot-filling used to generate conversations between a user and system bot, which are then rewritten and validated by crowdworkers. 
Thus, this still involves a considerable amount of human oversight and involvement.

\paragraph{LLMs as Agents.} The quality of dialogue and reasoning in current LLMs also has enabled research on simulating human agents: For instance, \citet{argyle2023out} demonstrate that suitable prompts can make LLMs mimic the answers of different test subject demographics, and \citet{liu2022generative} create a dialogue user-simulator. \citet{liu2023training} put LLM agents equipped with memory into a grid world and stimulate discussions between them about sensitive topics to aid the emergence of social norms. By further having LLMs act with different personas and enabling them to retrieving relevant memories, \citet{park2023generative} are able to simulate the social dynamics in a simulated village.

\paragraph{LLM Self-Improvement.} While na\"ively re-training a generative model on its own outputs might produce degenerate solutions \citep{alemohammad2023self, shumailov2023model}, LLMs have been successfully shown to produce data augmentations for other models \citep{yoo2021gpt3mix, dai2023chataug}. Furthermore, a recent line of work is exploring the limits of using LLMs to provide signal for their own training: \citet{bai2022constitutional} partially automate the process of ranking model responses through human feedback by having models critique their generations according to different harmfulness criteria, while \citet{scheurer2023training} have the LLM generate multiple refinements of the output based on the human feedback. A further improvement in responses has been observed through the use of an external critique model \citep{paul2023refiner}, or through self-critiquing for a diverse set of tasks such as information extraction \citep{gero2023self}, code generation \citep{chen2023teaching} and summarization \citep{saunders2022self}.
We see the recent work of \citet{gulcehre2023reinforced} as closest to ours; the authors show how after generating data during a \emph{grow} step, different tranches can be utilized to incrementally train the model during an \emph{improve step}. 
In contrast, our approach focuses on a single improve step and employs a conceptually simpler supervised finetuning strategy over a RL objective. 
Furthermore, \citeauthor{gulcehre2023reinforced} focus their work on conditional language modeling, while we experiment within the domain of task-oriented dialogue.
This also motivates our framework of using two separate models for the dialogue participants, albeit that it also admits the use of two copies of the same model (see next section).

\paragraph{Task-oriented Dialogue \& LLMs.} Lastly, the way we prompt our models to follow certain workflows is reminiscent of other works for task-oriented dialogue: \citet{li2022controllable} are able to simulate both agent and user through dynamically selected prompts. 
In \citet{zhang2023sgp}, the authors guide LLMs through an external prompter module and generate queries to an external fact database that inform the final generation.
\citet{zhao2022anytod} propose a programmable framework for dialogue flows which supplies the model with program logic, a corresponding ontology and previous conversational states for a reply. In comparison, our model does not rely on external models to achieve a certain behavior.\\

We take these research lines as encouraging pieces evidence for the following points: a) Neural networks, and particular conversational agents, can be used to bootstrap their own training data under certain conditions and b) LLMs have the ability to provide learning signals for themselves. These insights are synthesized into our proposed method.

\section{Method}

We first introduce some notation for our method. 
We start with two parameter vectors $\btheta_\text{client}$ and $\btheta_\text{agent}$  of pretrained LLMs for the client and agent model, respectively. 
Since we might update the parameters multiple times, we denote $\btheta_\text{agent}^{(0)}$ / $\btheta_\text{client}^{(0)}$ as the original parameters and $\btheta_\text{agent}^{(t)}$ / $\btheta_\text{client}^{(t)}$ as the parameters at time step $t$ of our procedure. Next, we define two different prompts $\bpr_\text{agent}$ and $\bpr_\text{client}$ that supply character descriptions and behavioral instruction to both models.
These models together parameterize a distribution over possible dialogues $p(\bd | \btheta_\text{client}^{(t)}, \btheta_\text{agent}^{(t)}, \bpr_\text{agent}, \bpr_\text{client})$.\\

For our method in \cref{fig:schema}, we now sample $N$ dialogues $\mathcal{D}^{(t)} = \{\bd_1^{(t)}, \ldots, \bd_N^{(t)} \}$ from this distribution at every time step by having the models converse in roles specified by their prompts.
While we can expect some of the generated dialogues to be successful given a sufficient model and high-quality prompts, there is no guarantee for them to  achieve novel dialogue goals.
For this reason, we employ a filter function to create a subset $\mathcal{D}^{(t)}_\bigtriangledown \subseteq \mathcal{D}^{(t)}$ of the generated dialogues.
An example for a filter would be to only include successful conversations, identified using a method we introduce later in \cref{sec:evaluation}.
The filtered examples are used to finetune the models by minimizing the negative log-likelihood

\begin{equation}
    - \frac{1}{|\mathcal{D}^{(t)}_\bigtriangledown|}\sum_{\bd^{(t)} \in \mathcal{D}^{(t)}_\bigtriangledown}\log p_{(\cdot)}\big(\bd^{(t)} \big| \btheta_{(\cdot)}^{(t)}),
\end{equation}

\noindent where after finetuning, we obtain a new parameter set $\btheta_{(\cdot)}^{(t + 1)}$, with which we can start the process again.
This framework allows many different variations: $\btheta_\text{agent}$ and $\btheta_\text{client}$ can be copies of the same or different model; we can finetune one or both copies. 
We use different models for agent and client, and do not update the client model $(\btheta_\text{client}^{(t)} = \btheta_\text{client}^{(0)}$).

\subsection{Prompt Design}\label{sec:turn-based-prompting}

\begin{figure}[tb!]
    \centering
    \includegraphics[width=0.985\columnwidth]{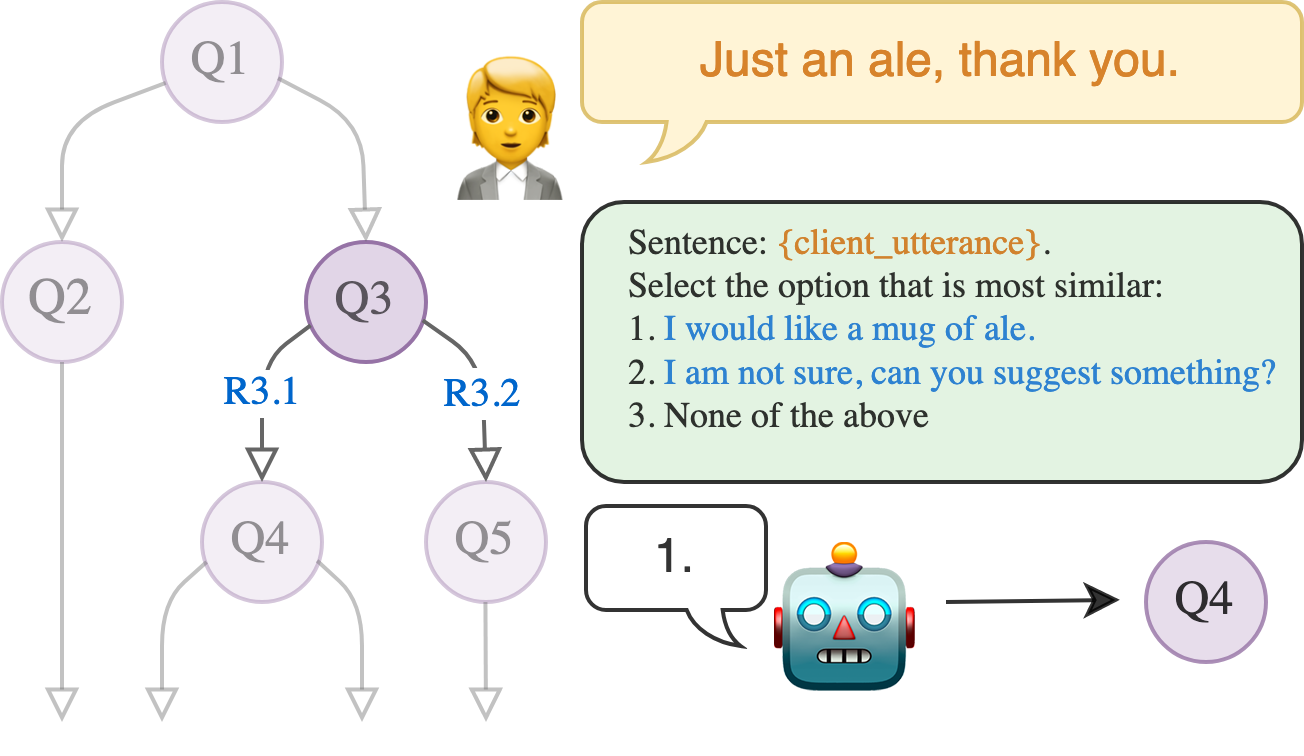}
    \caption{Illustration of the structured prompting: Workflows are parsed into a directed graph (left). At every turn of the conversation, we ask a LLM to compare the \textcolor{slotcolor}{client's last utterance} with the \textcolor{referencecolor}{reference responses} corresponding the outgoing edges of the current node. If one of them is chosen, we continue with the next node in the graph and prompt the agent with the corresponding question next turn, otherwise we stay in the same place and in graph and let the model generate freely.}\label{fig:model-prompting}
\end{figure}

\noindent In order to obtain both successful conversations for finetuning that are also sufficiently diverse, we dedicate special care to the design of the prompts $\bpr_\text{agent}$ and $\bpr_\text{client}$, which we lay out next.

\paragraph{Agent \& Client Prompts.} We create dialogues by prompting both client and agent with a character role, a persona, the dialogue history so far and the intention for the client as well as instructions for the agent.
The different combinations of characters, intentions and the stochastic nature of the generation thus creates a variety of scenarios. 
The prompts are described in detail in \cref{app:self-talk-look}. 

\paragraph{Structured Prompting.} 
In preliminary experiments, we found our models struggling to follow provided workflows, which we mostly attribute to model size and quality of the underlying pretraining data.
This negatively impacted the quality of bootstrapped conversations, providing a noisy training signal when finetuning the agent model.
To mitigate this issue, we parse workflows into directed graphs $\mathcal{G} = \{\mathcal{V}, \mathcal{E}\}$, in which vertices $\mathcal{V}$ correspond to questions of the agent and edges $\mathcal{E}$ to pre-defined reference answers of the client. 
Based on these, we propose \emph{structured prompting}, which we show in \cref{fig:model-prompting}:
At every step of the conversation, we consider the node of the last agent utterance $\bv_i \in \mathcal{V}$ and
provide a LLM with a set of suggested client responses that correspond to $\bv_i$'s outgoing edges and ask it to select the option that comes closest to the client's actual response (including a ``None of the above'' option). 
We then prompt the agent with the next question that follows from this response before the next turn and move up in the graph to the next node $\bv_\text{next} \in \text{Children}(\bv_i)$ until the conversation ends or a leaf node in the graph is reached. 
We show the full prompt in \cref{app:self-talk-look}.

\section{Experiments}\label{sec:experiments}

\paragraph{Dataset.} For the characters that the LLMs are supposed to embody in our conversations, we source descriptions from the LIGHT dataset by \citep{urbanek2019learning}.
Originally used for agents that solve tasks in a fantasy-world environment, it contains different 7487 quest descriptions, including characters and their personas. 
For instance, the role of \texttt{genie from the lamp} has the persona of \emph{``I am a genie and I enjoy playing tricks on those who release me from my lamp. I was imprisoned in this lamp by an evil witch. I am very bitter that I have to spend my days as a prisoner''}.
We choose a set of five characters for potential roles of the agent (\texttt{shop keeper}, \texttt{doctor}, \texttt{witch}, \texttt{tavern owner}, \texttt{genie from lamp}) and select another set of the $16$ most frequently occurring characters as potential clients (see \cref{app:dataset-creation}).
To create dialogue workflows, we then then prompt a 13 billion parameter OpenLlama variant \citep{touvron2023llama, together2023redpajama, openlm2023openllama}\footnote{Available under \url{https://huggingface.co/VMware/open-llama-13b-open-instruct}.} to generate three distinct workflows per agent character respectively including four, six and seven questions, which are then manually edited for quality. 
An overview and sample of these workflows is presented in \cref{app:dataset-creation}.

\paragraph{Self-talk Loop.} For the self-talk loop in \cref{fig:schema}, we choose the 30 billion parameter MosaicAI chat model \citep{MosaicML2023Introducing} for the client and the 7 billion parameter model from the same family for the agent. 
We found the client model to already work quite consistently using only prompting, so we choose to only finetune the agent model.
For every dialogue generation round, we first sample at random a client character, and agent character and a corresponding agent workflow.
Then, we generate the conversation utterance by utterance for up to eight turns (one turn including one utterance per character).
We also check whether a conversation has ended by using a combination of string matching and prompting a LLM along with some postprocessing to ensure the quality of generation, all of which are described in \cref{app:self-talk-look}. 

\paragraph{Filters.} Generated dialogues are then filtered and used to finetune the agent model for a single epoch. 
We compare the impact of multiple different filtering procedures, with baselines using the entire generated set of dialogues for finetuning (\textsc{None}) and sampling a random $p$ \% subset ($p$-\textsc{Random}).
We also explore whether at least $k$ subgoals or $p$ \% of subgoals in the workflow have been achieved ($k$-\textsc{Subgoals} and $p$-\textsc{Subgoals}, see next section).

\subsection{Evaluation}\label{sec:evaluation}

In order to assess the effectiveness of our approach, we evaluate it across the following axes.


\paragraph{Dialogue Diversity.} We measure the diversity of dialogues by tracking the number of unique words (\# \textsc{Words}) and unique $n$-grams (\#$n$-\textsc{Grams}; up to order 5).
We also measure diversity by measuring the average pair-wise ROUGE-L between $25$ dialogues of the same character (\textsc{Diversity}).

\paragraph{Subgoal Completion.} But what makes a successful conversation? Ideally, the agent should be guided through the dialogue flows towards one of the possible conversation endings and achieve their initial objective (see \cref{app:dataset-creation} for some examples). 
\citet{urbanek2019learning} define quests in the form of directed graphs that are traversed by the agents choosing actions in the environment from a large but finite set. 
Since potential actions are in text generation are infinite, the decision of whether a certain step in the conversation has been reached is much harder to decide.
For this reason, we use the same setup as in \cref{sec:turn-based-prompting} / \cref{fig:model-prompting} to parse our dialogue flows into directed graphs as well, where every node is associated with an utterance by the agent.
To decide whether to proceed along an edge, we compare whether the actual utterance of the agent and the reference utterance in the dialogue flow are similar enough.
We measure this similarity by means of ROUGE-L \citep{lin2004rouge} as implemented by the \texttt{evaluate} package,\footnote{See \url{https://github.com/huggingface/evaluate}.} which is then compared against pre-defined thresholds.
We explain how the initial thresholds are set on an external dataset in \cref{app:automatic-evaluation} and validate the soundness of these metrics at the end of this section.
For evaluation purposes, we track whether the dialogue reached one of its possible endings (\textsc{Dialogue Success}) and the absolute number of fulfilled workflow steps as measured by the depth in the directed graph (\textsc{Abs. Depth}) and as a proportion of the maximum depth (\textsc{Rel. Depth}), so intuitively the relative completion of the workflow.

\paragraph{Character Consistency.} Another questions regards whether the client stays in character or ignores the prompt and generates irrelevant outputs.
We take inspiration from the dungeon master model by \citet{shuster2020deploying} that was proposed to encourage human participants to role-play in dialogues.
For that purpose, we finetune the DeBERTa v3 model \citep{he2021debertav3, he2021deberta} on $200$k positive and $200$k negative examples  using a simple binary cross-entropy objective. 
The LIGHT dataset contains episodes with dialogues corresponding to quests, and we use those to create positive examples by sampling an episode and grouping the character name and persona with a random utterance from the conversation.
For negative examples, we sample a random utterance by a different character from another conversation conversation.
In addition, we add all steps of our workflows to the training data and upsample them by a factor of $308$, resulting in around $500k$ training samples in total.
More information on the training is given in \cref{app:automatic-evaluation}.
In our experiments, we evaluate the character consistency by scoring each utterance using the dungeon master model and averaging the scores over the entire conversation (\textsc{Char. Cons.}).\\

\paragraph{Validation of Automated Metrics.} Before we continue to our experiments, we first would like to validate our automatic evaluation metrics. 
For this purpose, we sample $50$ conversations from the untrained model and manually annotate them, before comparing the results of the automated metrics to the ground truth. 
We find a Kendall's $\tau$ correlation of $0.67$ for absolute depth and a Pearons's $\rho$ correlation of $0.67$ for relative depth for the subgoal tracking.
The dungeon master model showed an accuracy of $0.87$ and $F_1$ score of $0.93$ on the conversations.
We provide a more extensive analysis of these results in 
\cref{app:meta-evaluation}.

\subsection{Bootstrapping Finetuning Data}\label{sec:bootstrapping-experiment}

\begin{table*}
    \centering
    \caption{Results of the single step loop using different filters. 
    Shown are the percentage of filtered dialogues (\% \textsc{Filtered}), agent chararacter consistency (\textsc{Char. Cons.}), number of completed subgoals ($\varnothing$ \textsc{Abs. Depth}) and relative workflow completion  ($\varnothing$ \textsc{Rel. Depth}), percentage of finished conversations ($\%$ \textsc{Ended}), as well as dialogue diversity. Results stem from $100$ test dialogues and random filter scores from $3$ different seeds.}\label{tab:results-experiment1}
    \resizebox{.995\textwidth}{!}{
    \renewcommand{\arraystretch}{1.6}
    \begin{tabular}{lrrrrrrrr}
        \toprule
        \textsc{Data} & \small  \% \textsc{Filtered} & \small  \textsc{Char. Cons.} $\uparrow$ & \small  $\varnothing$ \textsc{Abs. Depth} $\uparrow$ & \small  $\varnothing$ \textsc{Rel. Depth}  $\uparrow$ & \small  $\%$ \textsc{Ended}  $\uparrow$ & \small  \# \textsc{Words}   $\uparrow$ & \small  \# $N$-\textsc{Grams} $\uparrow$ & \small \textsc{Diversity} $\uparrow$ \\
        \midrule
         \small \textsc{Initial Dialogues} & - & $0.77$ & $2.15$ & $0.38$ & $0.26$ & $57.16$ & $373.87$ & $0.69$ \\[0.1cm]
         \cdashline{1-9}
         \small  \textsc{No Filter} & $0.00$ & $0.78$ & $2.22$ & $0.38$ & $0.29$ & $47.32$ & $285.94$ & $0.60$ \\
         \small  $\%$-\textsc{Random} ($0.1$) & $0.90$ & $\underset{\small \pm 0.00}{0.77}$ & $\underset{\small \pm 0.10}{2.15}$ & $\underset{\small \pm 0.02}{0.37}$ & $\underset{\small \pm 0.03}{0.24}$ & $\underset{\small \pm 0.66}{45.89}$ & $\underset{\small \pm 6.21}{273.71}$ & $\underset{\small \pm 0.01}{0.62}$ \\
         \small  \textsc{Abs. Subgoals} ($4$) & $0.82$ & $0.78$ & $2.30$ & $0.39$ & $0.27$ & $46.62$ & $297.78$ & $0.64$ \\
         \small  \textsc{Abs. Subgoals} ($5$) & $0.92$ & $0.80$ & $2.41$ & $0.41$ & $0.32$ & $46.55$ & $280.53$ & $0.63$ \\
         \small  \textsc{Abs. Subgoals} ($6$) & $0.98$ & $\mathbf{0.81}$ & $2.11$ & $0.35$ & $0.22$ & $54.14$ & $343.07$ & $0.64$ \\
         \small  $\%$-\textsc{Subgoals} ($0.1$)& $0.90$ & $0.76$ & $2.37$ & $0.39$ & $0.30$ & $44.91$ & $266.77$ & $0.63$ \\
         \small  $\%$-\textsc{Subgoals} ($0.05$) & $0.95$ & $0.78$ & $\mathbf{2.54}$ & $\mathbf{0.43}$ & $\mathbf{0.36}$ & $49.40$ & $303.65$ & $0.66$ \\
         \small  $\%$-\textsc{Subgoals} ($0.01$) & $0.99$ & $0.78$ & $2.31$ & $0.39$ & $0.29$ & $\mathbf{55.25}$ & $\mathbf{355.44}$ & $\mathbf{0.69}$  \\
         \small  \textsc{Has Ended} & $0.74$ & $0.76$ & $2.22$ & $0.38$ & $0.26$ & $45.51$ & $277.25$ & $0.64$ \\
        \bottomrule
    \end{tabular}%
    }
\end{table*}

For our first experiment, we pre-generate around $9000$ dialogues to ensure comparability between different filters and finetune the model for a single epoch on the filtered data. 
We then generate an extra $100$ conversations with the updated agent model and measure the change in evaluation metrics.

\paragraph{Results.} Result for different filtering methods are shown in \cref{tab:results-experiment1}.
We can observe that using no or a random filter produces no or minor improvements at best. 
The best results are achieved by filters that select dialogues with at least $5$ completed workflow steps or the best $5 \%$ of dialogues in terms of completion, with drops in performance once these criteria are made stricter. 
This is mostly likely an effect of the shrinking finetuning set size, as only $1 - 2 \%$ of dialogues remain. 
Interestingly, dialogue diversity decreases in all cases; we attribute this to the fact that especially an agent model that sticks to a given workflow and does not go off-script creates dialogues that are overall slightly less diverse.

\begin{figure}[tb!]
    \begin{subfigure}[t]{\columnwidth}
        \centering
        \includegraphics[width=0.925\columnwidth]{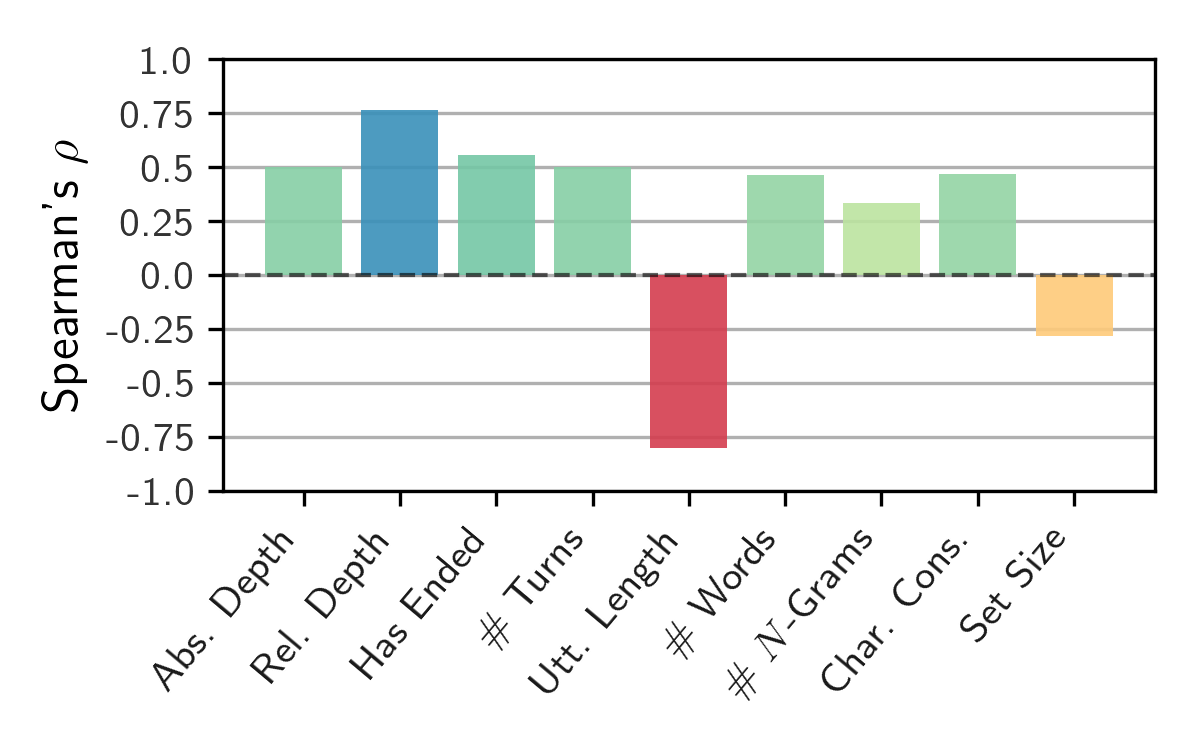}
        \caption{Importance via Spearman's $\rho$ correlation.}
        \label{subfig:importance-correlations}
    \end{subfigure}
    \begin{subfigure}[t]{\columnwidth}
        \centering
        \includegraphics[width=0.925\columnwidth]{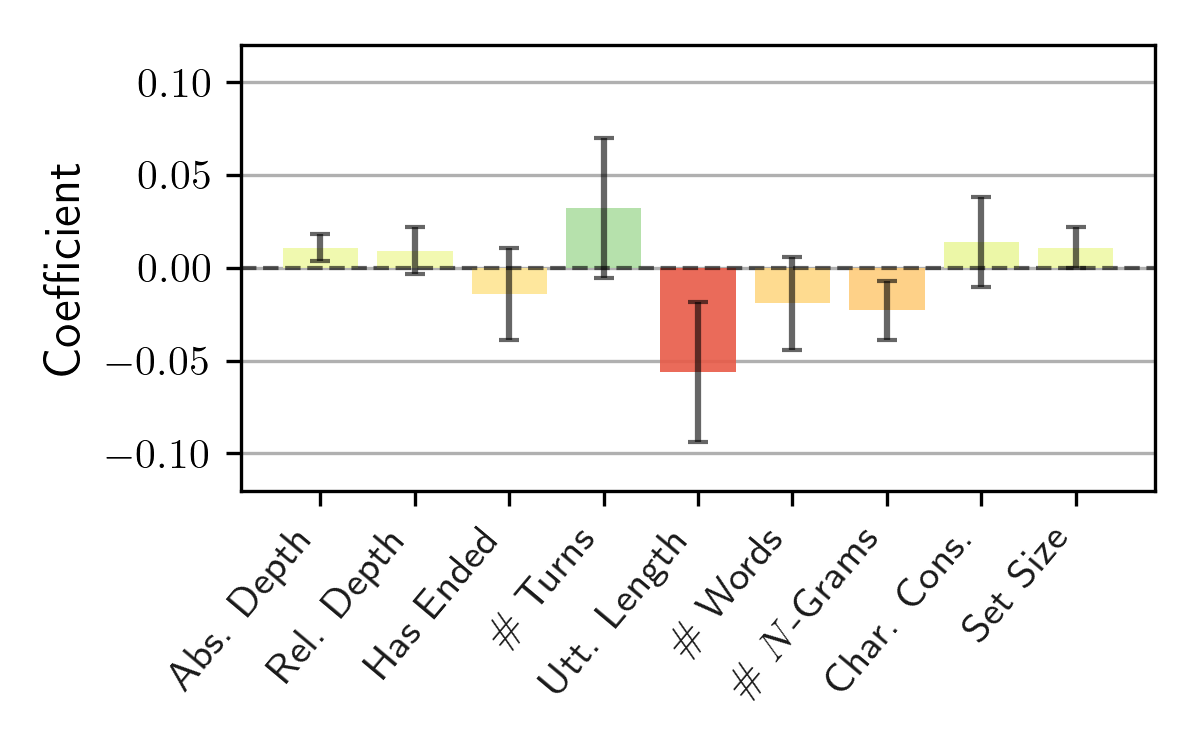}
        \caption{Importance via linear regression coefficients.}
        \label{subfig:importance-coefficients}
    \end{subfigure}
    \caption{Analysis of the relationship between properties of the finetuning dataset and their impact on the absolute completion of the dialogue, given (a) Spearman's $\rho$ correlation values and (b) the coefficients of the linear regression model without a bias and with lasso regularization. Error bars and the regularization weight were determined via cross-validation.}\label{fig:feature-importance}
\end{figure}

\paragraph{What Makes Good Finetuning Data?} The results in \cref{tab:results-experiment1} warrant the question for what reason different filtered finetuning sets worked better than others. 
We try to answer this by again applying our automated evaluation metrics---this time exclusively on the filtered dialogues---and relating the results to the observed change in dialogue success, measured in absolute depth.
We do this by collecting the properties of each finetuning set used in \cref{tab:results-experiment1} and running two different analyses: Correlating with them with the change in absolute depth in \cref{subfig:importance-correlations} using Spearman's $\rho$ and applying normal scaling to the features and then fitting a linear regression model (without bias) with $l_1$ regularization in \cref{subfig:importance-coefficients}.
We can see that the two methods agree on the fact that the number of turns has a positive effect on performance, while the average length of utterances impacts dialogue quality negatively.
This coincides with the intuitions we gained from our experiments, where bad quality conversations had models produce long utterances by rambling, going off-scrip, stating unnecessary information and ending prematurely or being shorter than the underlying dialogue workflow would require them to. 
Other than those, the methods agree that the workflow progression and character consistency in the finetuning data have a positive impact, with some slightly contradictory results for the other properties shown. 
However, these analyses should be taken with a grain of salt due to the small number of observations.
Also, it is unlikely that e.g.\@ utterance length has a negative effect on agent improvement \emph{per se}, it seems more plausible that all or many of the properties are influenced by a common, latent confounder, (namely dialogue quality), that manifests itself in our data along these axes.
Therefore, while we can use these observations for a proxy of the usefulness of a finetuning set, they do not imply causality. 

\subsection{Human Evaluation}\label{sec:human-evaluation}

\begin{figure*}[tb!]
    \centering
    \includegraphics[width=0.995\linewidth]{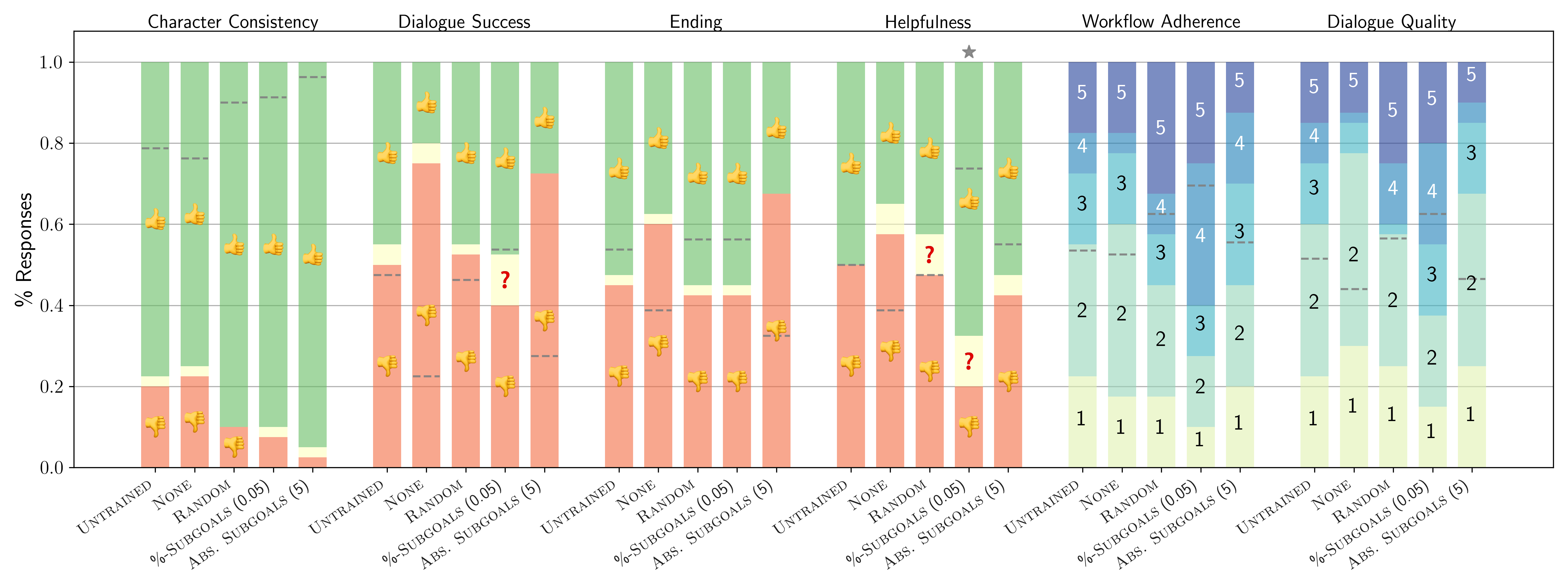}
    \caption{Results of the human evaluation study for three baselines and the two best filters from \cref{sec:bootstrapping-experiment} along six different questions. Shown are the percentage of ratings per filter, either on a five point scale or using positive (\thumbsup), negative (\thumbsdown) and \emph{unsure}  (\questionmark) options. Dashed lines indicate the numerical average and $\bigstar$ signifies statistical significance compared to all other options assessed via the ASO test \citep{del2018optimal, dror2019deep, ulmer2022deep} with $\tau = 0.5$ and a confidence level of $\alpha = 0.9$.}\label{fig:human-evaluation}
\end{figure*}

In order to triangulate the previous results that are based on automatic results and to additionally rate other aspects, we also include an evaluation based on human judges.
We evaluate the results of the untrained model (\text{Untrained}), using the entire finetuning data (\textsc{No Filter}), \textsc{\%-Random (0.1)}, \textsc{Abs. Subgoals (5)} and \textsc{\%-Subgoals (0.05)} from \cref{sec:bootstrapping-experiment} across certain dimensions such as helpfulness, consistency and dialogue success.
Users are asked to rate these properties either using three options (\thumbsup, \thumbsdown\ or \emph{unsure}) or on a scale from $1$ (worst) to $5$ (best).
We ask $n=9$ participants to judge a total of $100$ dialogues ($20$ per filtering method) randomly sampled from our experiments. 
Obtaining two annotations per data point, we observe an inter-annotator agreement in terms of Cohen's $\kappa \approx 0.52$.
We give more details about our evaluation including annotation guidelines and information on our tool in \cref{app:human-evaluation-study}.

\begin{figure}[tb!]
    \centering
    \begin{tcolorbox}[width=\columnwidth,colback={textbgcolor},title={{\small ``Make the prince fall in love with them''}}, enhanced,attach boxed title to top right={yshift=-3.5mm, xshift=-5mm}, colbacktitle=white, coltitle=black, top=12pt]   
        \small
        \agent{Genie from lamp}{Good day, what do you wish for?}
        \client{Villager}{Greetings, genie. I wish for the prince to fall in love with me.}
        \agent{Genie from lamp}{I am not a matchmaker and cannot make people fall in love. You should try to make your own chances.}
        \client{Villager}{Thank you for your advice, genie. I will try my best to make the prince fall in love with me.}
        \agent{Genie from lamp}{Goodbye!}
        \client{Villager}{Goodbye, genie.}
    \end{tcolorbox} 
    \caption{Conversation generated after finetuning with the \%-\textsc{Subgoals (0.05)} filter, with the agent ignoring the given workflow.}
    \label{fig:example-workflow-ignore}
\end{figure}

\begin{figure}[tb!]
    \centering
    \begin{tcolorbox}[width=\columnwidth,colback={textbgcolor},title={{\small ``Book a bed for the night''}}, enhanced,attach boxed title to top right={yshift=-3.5mm, xshift=-5mm}, colbacktitle=white, coltitle=black, top=12pt]   
        \small
        \agent{Tavern Owner}{Good day, how can I help you?}
        \client{Person}{Good day, I am in need of a bed for the night. I have been traveling for days and am quite tired.}
        \agent{Tavern Owner}{What is the duration of your stay?}
        \client{Person}{Just for the night, I have to continue my journey in the morning.}
        \agent{Tavern Owner}{What kind of room would you like?}
        \client{Person}{A simple room with a bed and a fireplace would be perfect.}
        \agent{Tavern Owner}{Good day, how can I help you?}
        {\hspace{3cm} ...}\\
        \textcolor{referencecolor}{[Agent gets stuck in a loop, asking the last two questions over and over.]}
    \end{tcolorbox} 
    \caption{Conversation generated after finetuning with the \textsc{Abs. Subgoals (5)} filter, with the agent getting stuck in a conversational loop.}
    \label{fig:example-conversation-loop}
\end{figure}

\paragraph{Results} The results of the aggregated judgements by filter are given in \cref{fig:human-evaluation}.
Interestingly, we can see that finetuning on all data usually \emph{decreases} model quality, but finetuning on a random subset \emph{may} increase performance.
Compared to the baselines, the human judgements surprisingly rate the quality of the model produced by the \textsc{Abs. Subbgoals (5)} filter as en par or worse on dimensions such as dialogue success, workflow adherence and overall dialogue quality.
With the other best performing filter however, $\%$-\textsc{Subgoals}($0.05$), our human raters triangulate the positive results of the automated metrics, with the resulting agent performing (significantly) better in all but one category, showing that right subset of dialogues can provide a stronger training signal.
Despite these successes, we still identify some common failure cases, of which we highlight two here:\footnote{We use \textcolor{agentcolor}{Agent} \raisebox{-0.05cm}{\includegraphics[width=0.4cm]{img/agent.png}} to illustrate the utterances of agent characters (inlcuding the ones that do not correspond to the genie) and \textcolor{clientcolor}{Client} \raisebox{-0.05cm}{\includegraphics[width=0.4cm]{img/client.png}} for client characters.} 
In \cref{fig:example-workflow-ignore}, we recapitulate a conversation in which the agent completely ignores the workflow after its first utterance, even though the rest of the conversation still continues organically. 
In the second example in \cref{fig:example-conversation-loop}, we see that the agent regresses to the initial question after an otherwise successful start of the conversation, getting stuck in a conversational loop with the client.
We also found conversations to derail in cases where the client would respond in a way too dissimilar from the sample answers defined in the workflow; in this case the agent would usually ad-lib and fail to recover during the rest of the conversation. 
We show more examples and discuss other failure cases in \cref{app:sample-conversations}.

\section{Discussion}\label{sec:discussion}

Our experimental results have demonstrated the challenges that arise when implementing self-talk in a task-oriented dialogue setup. 
These reach from the ability of the agent to follow the workflow carefully and selecting the next steps reliably---a problem that is at least partially rooted in the open-ended and paraphrastic nature of conversations.
Nevertheless, we have provided insights that a proper filtering technique isolating high-quality data contitutes a positive training signal for the agent LLM, that can be used to improve its performance.
This conclusion is in line with the emerging literature on LLM self-critique and the concurrent work of \citet{gulcehre2023reinforced} on using self-generated data for finetuning. 
An open question in this vein is whether this process can reliably repeated multiple times, and determining the conditions under which it leads to an improvement or model degradation.
In some tentative experiments, we found multi-loops of self-talk to lead to minor improvements at first, but also to be unstable if the number of filtered finetuning samples was too low. 
We leave further explorations of this question to future work and collect other, smaller insights from our experiments in \cref{app:insights}.

\section{Conclusion}\label{sec:conclusion}

Despite previous works articulating worries about the distributional collapse of model trained on their own generated data \citep{alemohammad2023self, shumailov2023model}, we demonstrated that---given the right setup---language models can self-improve by being trained on their own generation and become better task-oriented dialogue agents.
As we showed in \cref{sec:bootstrapping-experiment}, this depends on the setup; using different models and varying prompts helps to create sufficiently diverse data, with additional filters ensuring sufficient quality. 
We especially found the filters that strike a balance between quality and number of total data points to work best for selecting finetuning data and validated these insights with the human evaluation in \cref{sec:human-evaluation}.\\
In addition, we established new methods for automatic evaluation of agents in a task-oriented dialogue setting and showed that they can be used to quantify the progress of our finetuned model. 
We see this work as a piece of evidence in line with works such as \citet{scheurer2023training, paul2023refiner, gero2023self, chen2023teaching, saunders2022self} that self-improvement with LLMs is indeed possible under the right conditions, and are---to the best of our knowledge---the first to demonstrate the utility of LLM self-talk in a dialogue setting.
We hope for this line of work to inspire further investigations into understanding the requirements and limits of self-improvement.

\paragraph{Future Work.} This work focused on self-talk for task-oriented dialogue agents. 
Since we are finetuning on bootstrapped conversations, we expect the model to loose some general conversational abilities that are not relevant for the given task.
Future work could explore how we can nevertheless maintain these abilities.
Futhermore, this version of self-talk could still be improved by generating higher-quality conversations and identifying the conversations are most informative to the model for finetuning, reminiscent of work in active learning (see e.g.\@ \citealp{zhang2022survey}).
Lastly, our setup intuitively lends itself well to also using negative signals, since low-quality conversation could be used for negative rewards.
The agent could then be optimized with an objective similar to direct preference optimization \citep{rafailov2023direct} or using natural language policy optimization \citep{ramamurthy2023is} by embedding it in a RL environment.

\section*{Limitations}

This work comes with some limitations: Firstly, we restrict our experiments to task-oriented dialogue, which does not allow us to draw any conclusion about open-ended dialogue or other, more specific tasks carried out by LLMs.
In order to isolate the effect of self-talk, we also limit the scope of the interaction between agent and client and constrain the format of the dialogue.
However, related works by \citet{scheurer2023training, paul2023refiner, gero2023self, chen2023teaching, saunders2022self} have demonstrated the potential for self-improvement for a variety of tasks.
Secondly, we explored a setup which two different language models that are further given very strict instructions in order to produce conversations that are usable for training.
It might be that using a single model for both the agent and the client might not produce conversation that are diverse enough, or that smaller models might not be able to create sufficiently useful conversations at all.
Furthermore, even including all the techniques like model-guided prompting the generated conversations sometimes would still include the models getting stuck in loops, going off-script or even producing nonsensical dialogues that can also be hard to catch with our defined filters sometimes.
Despite with anecdotal evidence that larger models might not suffer from these problem as much, it is hard to definitively disentangle the effect of model size and training data in our setup.

\section*{Ethical Considerations}

\paragraph{Model Bias.} Many works have been dedicated to investigating social biases of modern NLP models (see for instance \citealp{stanczak2021survey, garrido2021survey} for surveys).  
Since large language models often echo the biases present in their development \citep{waseem2021disembodied}, finetuning a model on its own bootstrapped data harbors the danger of amplifying its own biases.

\paragraph{Societal Dangers.} Recent works have demonstrated the negative potential of LLMs to deceive human users, for instance in the form of personalized phishing attacks \citep{hazell2023large}.
Training agents which do not rely on human-generated training data anymore could therefore simplify the creation of deceptive dialogue models by malicious actors.

\section*{Acknowledgements}

We would like to thank Wanyu Du and Han He for providing the inspiration for the model-guided prompting, and Nikolaos Pappas and Raphael Shu for their fruitful discussions about the work.
We also thank Elisa Bassignana, Sam Davidson, Etsuko Ishii,  Josef Valvoda, Yuwei Zhang and especially Hossein Aboutalebi for their efforts in contributing to our human evaluation study. 

\bibliography{anthology,custom}
\bibliographystyle{acl_natbib}

\appendix

\section{Implementation Details}\label{app:implementation-details}

This appendix section focuses on several implementation details important for replication, but which are too specific for the main text of this work. 
\cref{app:dataset-creation} is dedicated to the creation of the experimental data, while \cref{app:self-talk-look} mainly focuses on the implementation details of the self-talk loop and on the automatic evaluation in \cref{app:automatic-evaluation}.

\subsection{Data Creation}\label{app:dataset-creation}

\paragraph{Client Characters.} We select the following characters from the LIGHT dataset \citep{urbanek2019learning} to serve as potential clients: \texttt{king}, \texttt{person}, \texttt{guard}, \texttt{queen}, \texttt{peasant}, \texttt{child}, \texttt{knight}, \texttt{thief}, \texttt{bird}, \texttt{traveler}, \texttt{rat}, \texttt{villager}, \texttt{spider}, \texttt{ghost}, \texttt{horse}, and \texttt{farmer}.

\begin{table}
    \centering
    \caption{Overview over the used agent characters and their corresponding workflow topics, correspondings to four, six and seven total questions as listed.}\label{table:workflows-overview}
    \renewcommand{\arraystretch}{1.5}
    \resizebox{.975\columnwidth}{!}{
    \begin{tabular}{lr}
        \toprule
        \textsc{Agent Character} & \textsc{Workflow Topics} \\
        \midrule
        \texttt{shop keeper} & \makecell[r]{buy a longsword,\\ buy food for a long journey,\\ sell a valuable item} \\[0.75cm]
        \texttt{doctor} & \makecell[r]{ask how to treat a headache,\\ ask for a remedy for an fever,\\ ask how to treat an animal bite} \\[0.75cm]
        \texttt{witch} & \makecell[r]{give them a love potion,\\ curse their enemy,\\ give them a magic wand} \\[0.75cm]
        \texttt{tavern owner} & \makecell[r]{order a drink,\\ order a rich meal,\\ book a bed for the night} \\[0.75cm]
        \texttt{genie from lamp} & \makecell[r]{become rich,\\ make the prince fall in love with them,\\ become a ruler} \\
        \bottomrule
    \end{tabular}%
    }
\end{table}

\begin{figure*}
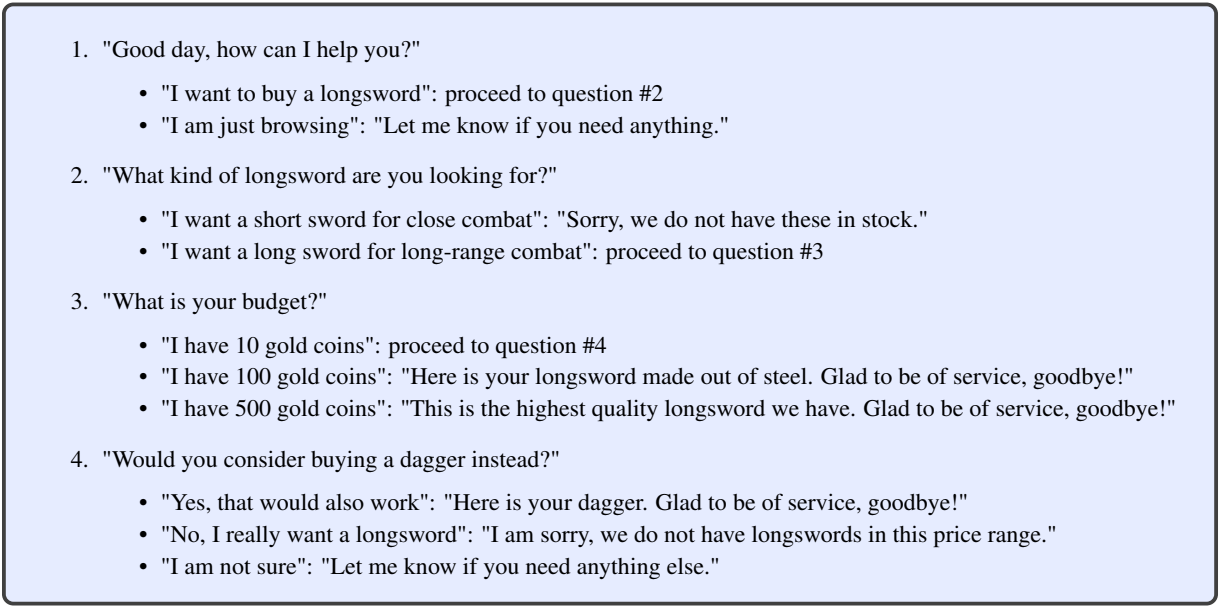

    \centering
    \begin{tcolorbox}[width=\textwidth,colback={workflowbgcolor},title={}] 
    \small
    \begin{enumerate}
        \item "Good day, how can I help you?"
        \begin{itemize}
            \item "I want to buy a longsword": proceed to question \#2
            \item "I am just browsing": "Let me know if you need anything."
        \end{itemize}
        \item[2.] "What kind of longsword are you looking for?"
        \begin{itemize}
            \item "I want a short sword for close combat": "Sorry, we do not have these in stock."
            \item "I want a long sword for long-range combat": proceed to question \#3
        \end{itemize}
        
        \item[3.] "What is your budget?"
        \begin{itemize}
            \item "I have 10 gold coins": proceed to question \#4
            \item "I have 100 gold coins": "Here is your longsword made out of steel. Glad to be of service, goodbye!"
            \item "I have 500 gold coins": "This is the highest quality longsword we have. Glad to be of service, goodbye!"
        \end{itemize}
        
        \item[4.] "Would you consider buying a dagger instead?"
        \begin{itemize}
            \item "Yes, that would also work": "Here is your dagger. Glad to be of service, goodbye!"
            \item "No, I really want a longsword": "I am sorry, we do not have longswords in this price range."
            \item "I am not sure": "Let me know if you need anything else."
        \end{itemize}
    \end{enumerate}
    \end{tcolorbox}
    \caption{Example workflow for the agent character of the \texttt{shop keeper} - buy a longsword.}
    \label{fig:workflow-shopkeeper}
\end{figure*}

\paragraph{Creating Agent Workflows.} To create agent workflows, we prompt an instruction-finetuned version of OpenLlama to generate dialogue workflows including four, six and seven questions or subgoals.
These are then manually edited to ensure logical consistency and adherence to the given format. 
The topic of the workflow is tailored toward the character and listed in \cref{table:workflows-overview}.
We also list some examples for workflows in \cref{fig:workflow-shopkeeper,fig:workflow-doctor,fig:workflow-genie}.

\begin{figure*}
    \centering
    \begin{tcolorbox}[width=\textwidth,colback={workflowbgcolor},title={}]    
    \small
    \begin{enumerate}
    \item "Good day, how can I help you?"
    \begin{itemize}
        \item "I was bitten by an animal": proceed to question \#3
        \item "I was just in the woods and got bitten by an insect": proceed to question \#2
    \end{itemize}
    \item[2.] "Is your bite swollen?"
    \begin{itemize}
        \item "Yes, it is swollen": "Here is some ointment for your swollen bite. Glad to be of service, goodbye!"
        \item "No, it is not swollen": "Come back tomorrow if anything changes. Glad to be of service, goodbye!"
    \end{itemize}
    
    \item[3.] "How is the wound?"
    \begin{itemize}
        \item "It is bleeding heavily": "Here is a bandage for your bleeding wound, come back tomorrow. Glad to be of service, goodbye!"
        \item "It is not bleeding heavily": proceed to question \#4
        \item "It is not bleeding and the wound looks clean": proceed to question \#5
    \end{itemize}
    
    \item[4.] "Has the wound been cleaned?"
    \begin{itemize}
        \item "Yes, I cleaned it with water": "Here is some alcohol to clean the wound. Come back tomorrow if anything changes. Glad to be of service, goodbye!"
        \item "Yes, I cleaned it with alcohol": proceed to question \#5
        \item "No, I have not cleaned it": "Here is some alcohol to clean the wound. Come back tomorrow if anything changes. Glad to be of service, goodbye!"
    \end{itemize}
    
    \item[5.] "What is the animal that bit you?"
    \begin{itemize}
        \item "I don't know": proceed to question \#6
        \item "It was a wolf": proceed to question \#6
        \item "It was a dog": proceed to question \#6
        \item "It was a bear": "Here is some medication for your pain. Come back tomorrow if anything changes. Glad to be of service, goodbye!"
    \end{itemize}
    
    \item[6.] "Do you have a fever?"
    \begin{itemize}
        \item "Yes, I have a fever": proceed to question \#7
        \item "No, I don't have a fever": "Come back tomorrow if anything changes. Glad to be of service, goodbye!"
    \end{itemize}
    
    \item[7.] "Do you have any other symptoms?"
    \begin{itemize}
        \item "The bite hurts a lot": "Here is some medication for your pain. Come back tomorrow if anything changes. Glad to be of service, goodbye!"
        \item "I feel dizzy": "Make sure to drink water, eat well and rest. Come back tomorrow if anything changes. Glad to be of service, goodbye!"
        \item "No, I don't have any other symptoms": "Come back tomorrow if anything changes. Glad to be of service, goodbye!"
    \end{itemize}
\end{enumerate}
    \end{tcolorbox}
    \caption{Example workflow for the agent character of the \texttt{doctor} - as how to treat an animal bite.}
    \label{fig:workflow-doctor}
\end{figure*}

\begin{figure*}
    \centering
    \begin{tcolorbox}[width=\textwidth,colback={workflowbgcolor},title={}]    
        \small
        \begin{enumerate}
        \item "Good day, what do you wish for?"
        \begin{itemize}
            \item "I want to make the prince fall in love with me": proceed to question \#4
            \item "I want to make the prince notice me": proceed to question \#2
            \item "I don't know": "Let me know how I can help you."
        \end{itemize}
        \item[2.] "Do you want the prince to fall in love with you?"
        \begin{itemize}
            \item "Yes, I want the prince to fall in love with me": proceed to question \#4
            \item "No, I just want the prince to notice me": proceed to question \#3
        \end{itemize}
        
        \item[3.] "In what setting do you want the prince to notice you?"
        \begin{itemize}
            \item "In a ballroom": "Here is your new outfit and invitation to the royal ball. Glad to be of service, goodbye!"
            \item "In a garden": "Here is your new outfit and invitation to the royal garden party. Glad to be of service, goodbye!"
            \item "In a royal banquet": "Here is your new outfit and invitation to the royal banquet. Glad to be of service, goodbye!"
        \end{itemize}
        
        \item[4.] "What is your reason for wanting the prince to fall in love with you?"
        \begin{itemize}
            \item "I want to have a lot of wealth": proceed to question \#5
            \item "I want to have a lot of power": proceed to question \#6
            \item "I want the beautiful prince to be my husband": "The prince is now in love with you and will seek you out soon. Glad to be of service, goodbye!"
        \end{itemize}
        
        \item[5.] "Do you just want me to make you rich instead?"
        \begin{itemize}
            \item "Yes, I just want to be rich": "Here is your new wealth. Glad to be of service, goodbye!"
            \item "No, I want the prince to fall in love with me": "The prince is now in love with you and will seek you out soon. Glad to be of service, goodbye!"
        \end{itemize}
        
        \item[6.] "Do you just want me to make you powerful instead?"
        \begin{itemize}
            \item "Yes, I just want to be powerful": "You are now the ruler of the kingdom. Glad to be of service, goodbye!"
            \item "No, I want the prince to fall in love with me": "The prince is now in love with you and will seek you out soon. Glad to be of service, goodbye!"
        \end{itemize}
    \end{enumerate}
    \end{tcolorbox} 
    \caption{Example workflow for the agent character of the \texttt{genie from lamp} - make the prince fall in love with them.}
    \label{fig:workflow-genie}
\end{figure*}

\subsection{Self-talk Loop}\label{app:self-talk-look}

In this section we give more information on the core ingredients of the self-talk loop, including the actual dialogue simulation, model prompts and finetuning details.

\paragraph{Dialogue Simulation.} The dialogue simulation follows these steps: An agent utterance and a client utterance, both followed by some postprocessing, then followed up by  a routine checking whether the conversation has ended or not.
For the generation, we sample up to 100 tokens using nucleus sampling with $p = 0.95$ \citep{holtzman2020curious} and top-$k$ sampling \citep{fan2018hierarchical, holtzmann2018learning, radford2019language} with $k=50$ and a temperature of $0.8$.
We also observed that models struggled to only generate a single utterance at a time, and would tend to try and generate the whole dialogue. 
Therefore, we implement a custom stopping criterion that stops sampling once the name of the other character is being generated.
For the postprocessing, we use regular expressions to remove the remaining fragments of this kind from the generation, along with incomplete sentences and code fragments.
The ending check includes checking for key phrases such as \texttt{good luck}, \texttt{goodbye}, \texttt{you're welcome} in the last two utterances. 
Furthermore, we send the last turn to the 30 billion parameter MosaicAI chat model using the following prompt:

\begin{tcolorbox}[width=\columnwidth,colback={promptbgcolor},title={\small End Detection Prompt},enhanced,attach boxed title to top right={yshift=-3.5mm, xshift=-5mm}, colbacktitle=white, coltitle=black, top=12pt]  
\small
You are an expert in identifying parts of a conversation.
This is an excerpt from a conversation:\\
\textcolor{slotcolor}{\{agent\_utterance\}}\\
\textcolor{slotcolor}{\{client\_utterance\}}\\

What part of a conversation is this from? 
Answer only [start/middle/end]. Correct answer:
\end{tcolorbox}

\noindent We give the LLM a single token for the answer, and finish the conversation when it is ``end''.
We furthermore speed up the generation process using the \texttt{accelerate} package\footnote{See \sloppy \url{https://huggingface.co/docs/accelerate/}} and always generate a batch of conversations at once, replacing single instances immediately when they finish.

\paragraph{Structured Prompting} In order to improve the degree of which the agent model follows a given workflow, we employ a turn-based prompting scheme. 
For this purpose, we parse every workflow into a directed graph with vertices corresponding to agent questions and edges to possible answers. 
Taking \cref{fig:workflow-shopkeeper} for instance, ``Good day, how I can help you?'' would correspond to the first node, and ``I want to buy a longsword'' and ``I am just browsing'' to outgoing edges.
Any question is given to the agent in the form of the following prompt: 

\begin{tcolorbox}[width=\columnwidth,colback={promptbgcolor},title={\small Agent Model Prompt},enhanced,attach boxed title to top right={yshift=-3.5mm, xshift=-5mm}, colbacktitle=white, coltitle=black, top=12pt]  
\small
\textcolor{specialtokencolor}{<|im\_start|>}\\
You are playing a \textcolor{slotcolor}{\{character\}}.\\ 
\textcolor{slotcolor}{\{first\_person\_persona\}}\\
Once and only if the conversation has ended, say "goodbye".\\
\textcolor{slotcolor}{\{dialogue\_history\_slot\}}\\
Now, you must say \textcolor{slotcolor}{\{turn\_prompt\_slot\}}.\\
\textcolor{specialtokencolor}{<|im\_end|>}
\end{tcolorbox}

\noindent where \texttt{first\_person\_persona} corresponds to a persona from the LIGHT datasets reworded in first person, \texttt{dialogue\_history\_slot} to the dialogue history up until that point (and empty in the beginning) and \texttt{turn\_prompt\_slot} to the current command, so ``Good day, how I can help you?'' in the beginning.
After the client respond to the agents initial questions, we then prompt the model with following:

\begin{tcolorbox}[width=\columnwidth,colback={promptbgcolor},title={\small Workflow Manager Prompt},enhanced,attach boxed title to top right={yshift=-3.5mm, xshift=-5mm}, colbacktitle=white, coltitle=black, top=12pt]  
\small
\textcolor{specialtokencolor}{<|im\_start|>}\\
You rate the similarity of texts.\\
Take the following sentence: \textcolor{slotcolor}{\{client\_utterance\}}.\\
Only select the option below that is most similar:\\ \\

\textcolor{slotcolor}{\{options\}}\\ \\

Correct answer:\\
\textcolor{specialtokencolor}{<|im\_end|>}\\
\end{tcolorbox}

\noindent where \texttt{options} contains all the outgoing edges as numbered options as well as a ``None of the above''. 
If one of the former options is selected by the model, we traverse the workflow graph along the corresponding edge and insert the next question into the first prompt.
If the ``None of the above'' option is selected, we remain in the same state and modify the last line of the first prompt to ``Now you must say any natural reply''.
We note that while it seem most intuitive to let the agent model choose the next options, we actually give this prompt to the client model, since it is larger and we can optimize it for inference since it is not finetuned. 
However, it would also possible to use the agent or even a third model for this purpose.

\paragraph{Client Prompting.} For completeness, we also share the prompt used for the client model, which is very similar in nature to the agent's prompt:

\begin{tcolorbox}[width=\columnwidth,colback={promptbgcolor},title={\small Client Model Prompt},enhanced,attach boxed title to top right={yshift=-3.5mm, xshift=-5mm}, colbacktitle=white, coltitle=black, top=12pt]  
\small
\textcolor{specialtokencolor}{<|im\_start|>}\\
You are playing a \textcolor{slotcolor}{\{character\}}. \textcolor{slotcolor}{\{first\_person\_persona\}}\\
Now, your main intention is to talk to the \textcolor{slotcolor}{\{agent\_character\}} to \textcolor{slotcolor}{\{intention\}}.\\
Once and only if the conversation has ended, tell them "goodbye".\\
Now, briefly respond to \textcolor{slotcolor}{\{agent\_character\}} and describe your intention.\\
\textcolor{specialtokencolor}{<|im\_end|>}\\
\textcolor{slotcolor}{\{dialogue\_history\_slot\}}
\end{tcolorbox}

\noindent In contrast to the agent, the client is given this prompt only in the beginning and then tries to continue the history of the conversation given in \texttt{dialogue\_history\_slot}. 
The client model is not being finetuned.

\paragraph{Finetuning.} Because it is infeasible to finetune the agent model in its entirety, we instead opt for the low-rank adapation or LoRA by \citet{hu2022lora}.
LoRA factorizes the gradient update $\Delta \mathbf{W} \in \mathbb{R}^{d \times d}$ for a parameter matrix $\mathbf{W} \in \mathbb{R}^{d \times d}$ into two low-rank matrices $\Delta \mathbf{W} = \mathbf{AB}$ with $\mathbf{A} \in \mathbb{R}^{d \times r}$ and $\mathbf{B} \in \mathbb{R}^{r \times d}$ s.t. $r \ll d$.
We apply these technique to all linear layers in the model using $r=64$ only optimize the these low-rank matrices with a learning rate of $5 \times 10^{-4}$, a weight decay of $0.01$ and batch size $4$ for a single epoch per finetuning loop.
All finetuning (as well as the dialogue generation from the previous step) is performed on four NVIDIA A10G Tensor Core GPUs.

\subsection{Automatic Evaluation}\label{app:automatic-evaluation}

In this section we give more details on the automated evaluation metrics we use to assess the quality of dialogues.

\paragraph{Subgoal Completion.} We lay out the method to measure subgoal completion in detail in \cref{sec:evaluation}.
In order to set the initial threshold thay ROUGE values have to surpass, we refer to another task-oriented dialogue dataset, namely ABCD \citep{chen2021action}. 
The dataset consists of $10042$ conversation with corresponding conversational workflows, of which we use the first $500$ in the following:
In order to match workflow steps to dialogue utterances, we assume that the correct utterance is the one that has the highest ROUGE-L score in comparison to the current workflow step.
We then collect all the previous utterances in the conversation as negative examples and use all the samples from the dataset to find the threshold that optimizes for precision, which we obtain as $0.33$. 
Even though optimizing for precision produces a more conversative threshold that is likely to underestimate subgoal completion, we found optimizing for recall to always yielding a trivial threshold of $0$, likely due to noise in the data.

\paragraph{Character Consistency.} We train our dungeon master using a batch size of $40$ for a total number of $15$ epochs. 
We also use the first $10 \%$ of the training as warmup for an inverse square-root learning rate schedule with a maximum learning rate of $5 \times 10^{-5}$.
To combat class imbalance, we compute class weights on the training set that are supplied to the loss function.
Training instances are supplied to the DeBERTA model using the format of \texttt{\textcolor{specialtokencolor}{[CLS]} \textcolor{slotcolor}{\{character\}}: \textcolor{slotcolor}{\{persona\}} \textcolor{specialtokencolor}{[SEP]} \textcolor{slotcolor}{\{utterance\}}} and context length of $256$ tokens.

\paragraph{Dialogue Diversity.} To measure the diversity in dialogues, we first try to quanity the number of unique words and $n$-grams (up to order $5$) in all dialogues corresponding to a specific agent character.
Furthermore, we compare up to $25$ pairs of dialogues per agent character and compute pair-wise ROUGE-L scores, of which we take the average. 
Since higher values for unique tokens and $n$-grams denote higher diversity, we define one minus the average pair-wise ROUGE-L score as a diversity metric.

\section{Experimental Appendix}\label{app:experimental-details}

This appendix supplies additional experimental analyses that had to be left out from the main paper due to spatial constraints.
\cref{app:meta-evaluation} gives further details on the meta-evaluation of our automated metrics laid out in \cref{sec:evaluation}, with \cref{app:human-evaluation-study} describing the human evaluation study from \cref{sec:human-evaluation}.
In addition, we discuss some miscellaenous insights gained from our work in \cref{app:insights} in the hope of aiding future research in this direction, while \cref{app:sample-conversations} displays a wide variety of sampled conversations, illustrating success stories as well as the most commonly observed failure cases.

\subsection{Validation of Automated Metrics}\label{app:meta-evaluation}

In the following we give more details for the validation of our automated metrics, 
from a description of the annotation guidelines (\cref{app:annotation-guidelines}), dialogue statistics (\cref{app:dialogue-statistics}) and finally the meta-evaluation of our automated metrics (\cref{app:meta-evaluation}).

\subsubsection{Annotation Guidelines}\label{app:annotation-guidelines}

In order to annotate our sampled conversations and contextualize subsequent results, we define the following annotation guidelines: 

\begin{itemize}
    \item \textsc{Character Consistency}: We take each utterance of the agent and client and score them with $0$ for out-of-character, $1$ for in-character and $0.5$ for undecided.
    The utterance is scored based on the character, persona and given intention.\footnote{Therefore, a bird wanting to become a king by asking a genie is still in character if that was the given intention.}
    We score the utterance in isolation from the wider dialogue context.
    \item \textsc{Subgoal Completion}: For every utterance of the agent, we mark it with $0$ if they do not correspond to a subgoal, and otherwise with \texttt{<question number>} or \texttt{<question number>.<agent response number>} when we end the conversation instead of proceeding to the next question.
    Taking for instance the workflow in  \cref{fig:workflow-shopkeeper}, $2.$ corresponds to the question \emph{``What kind of longsword are you looking for?''} and $4.3$ to \emph{``Let me know if you need anything else''}.
    Subgoals do not have to be completed in a given order.
    \item \textsc{Dialogue Success}: We score the dialogue with $1$ if the original intention of the client was fulfilled, $0$ in case it was not, and $0.5$ for undecided. This can be independent from whether the actual dialogue flow was adhered to or not.
    \item \textsc{Dialogue Quality}: We rate the quality of dialogues on a five point scale:
    \begin{itemize}
        \item \emph{1 - Very bad}: LLMs are not adhering to dialogue format, produce code or other unwanted artifacts.
        \item \emph{2 - Bad}: LLMs adhere to a dialogue format, but the dialogue contains loops, repetitions, major flaws in logic or LLMs break character or is very short.
        \item \emph{3 - Okay}: LLMs mostly stay in character and show only minor flaws in logic, dialogue flow and client intention are somewhat adhered to.
        \item \emph{4 - Good}: LLMs stay in character except for minor flaws in logic or role-playing, dialogue is successful and mostly follows the dialogue flow.
        \item \emph{5 - Very good}: LLMs always stay in character and show no flaws in logic or role-playing. The dialogue almost completely or completely follows the workflow.
    \end{itemize}
    \item \textsc{Ending}: $1$ if the conversation ended in a natural way, $0$ if not or if it didn’t end at all.
    This is to check whether the LLM-based ending detection described in \cref{app:self-talk-look} would work as expected.
\end{itemize}

\subsubsection{Dialogue statistics}\label{app:dialogue-statistics}

\begin{figure*}[tb]
    \centering
    \begin{subfigure}[t]{0.325\textwidth}
        \centering
        \includegraphics[width=0.985\columnwidth]{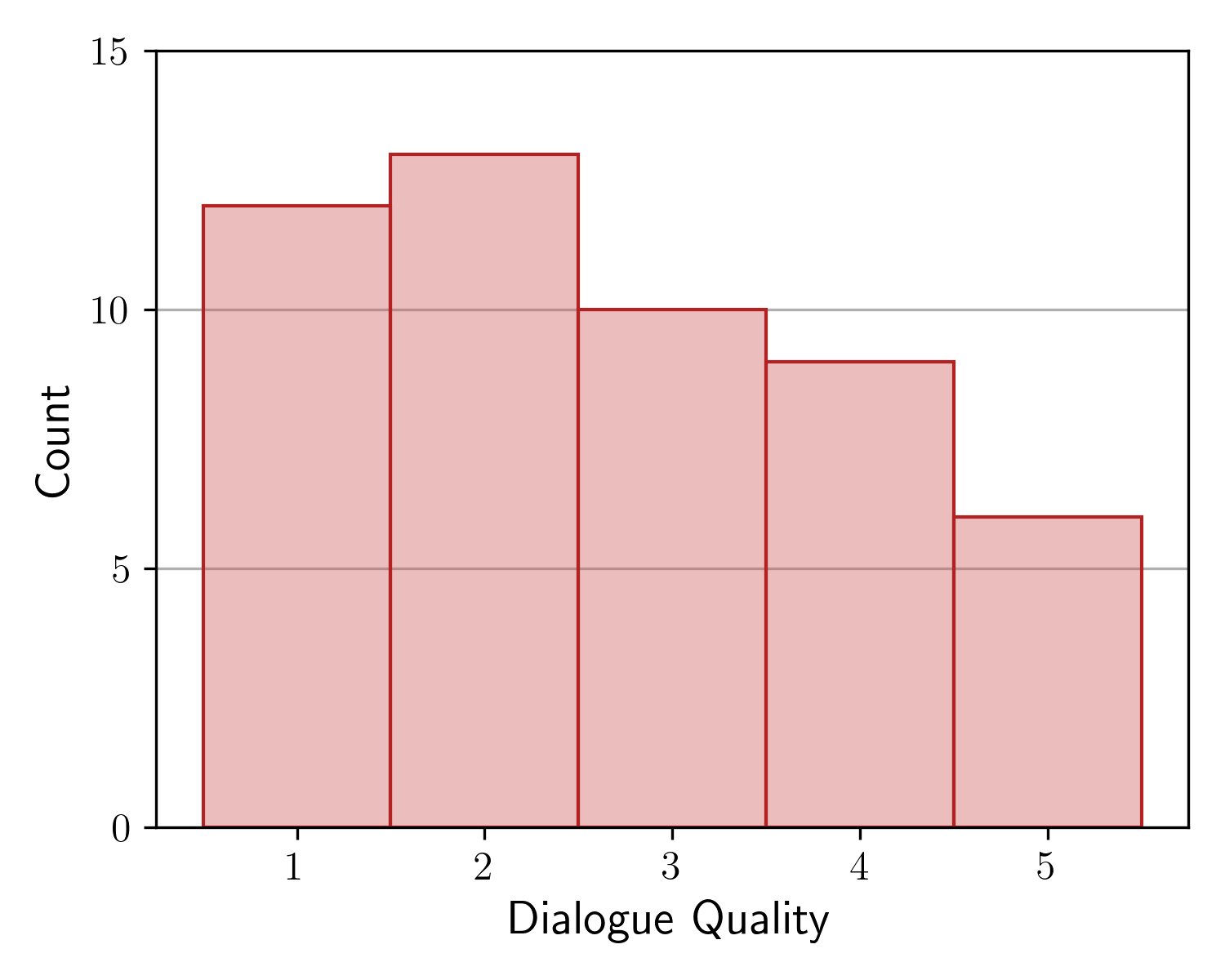}
        \caption{Dialogue quality.}
        \label{subfig:dialogue-quality}
    \end{subfigure}
    \hfill
    \begin{subfigure}[t]{0.325\textwidth}
        \centering
        \includegraphics[width=0.985\columnwidth]{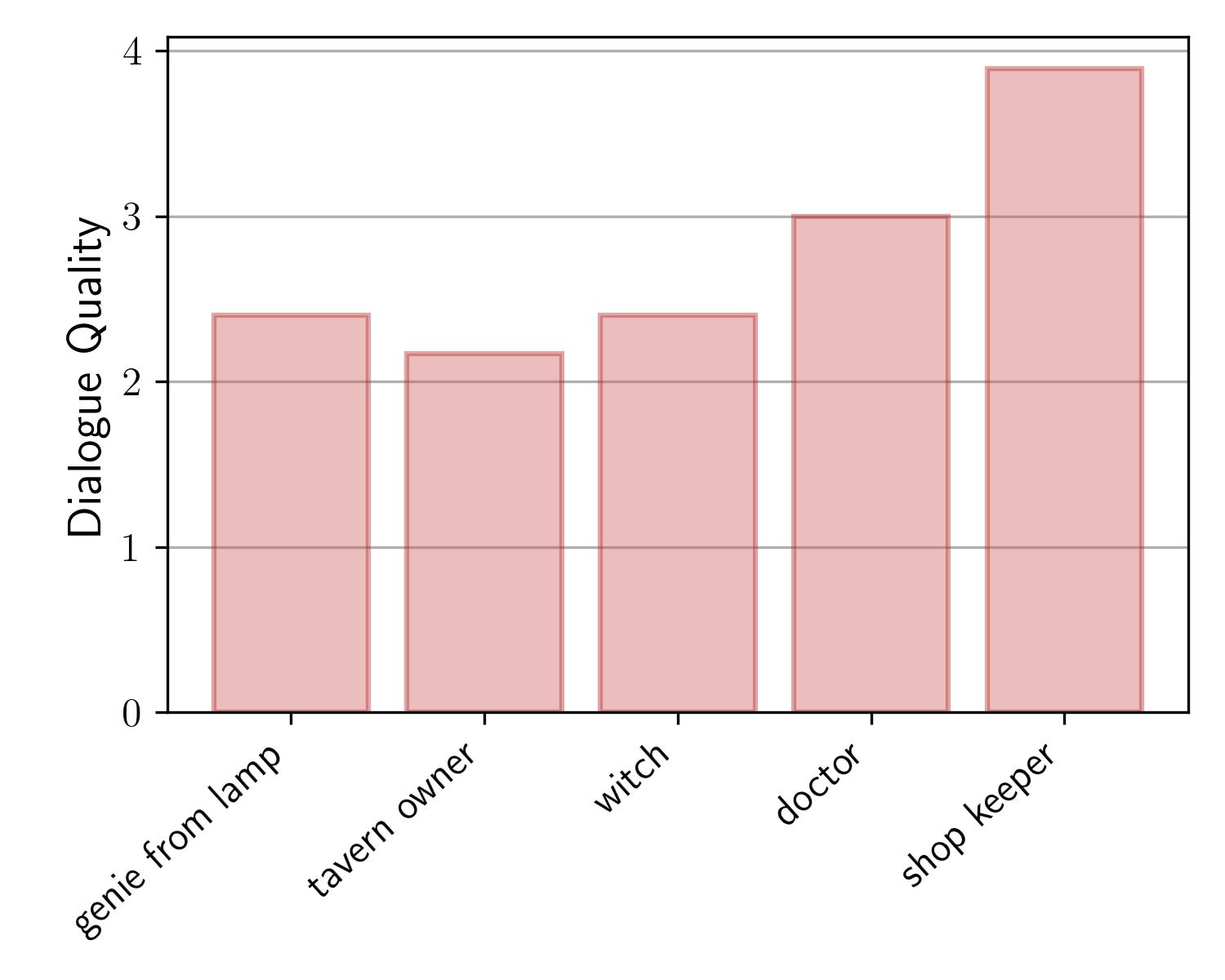}
        \caption{Dialogue quality by character.}
        \label{subfig:dialogue-quality-character}
    \end{subfigure}
    \hfill
    \begin{subfigure}[t]{0.325\textwidth}
        \centering
        \includegraphics[width=0.985\columnwidth]{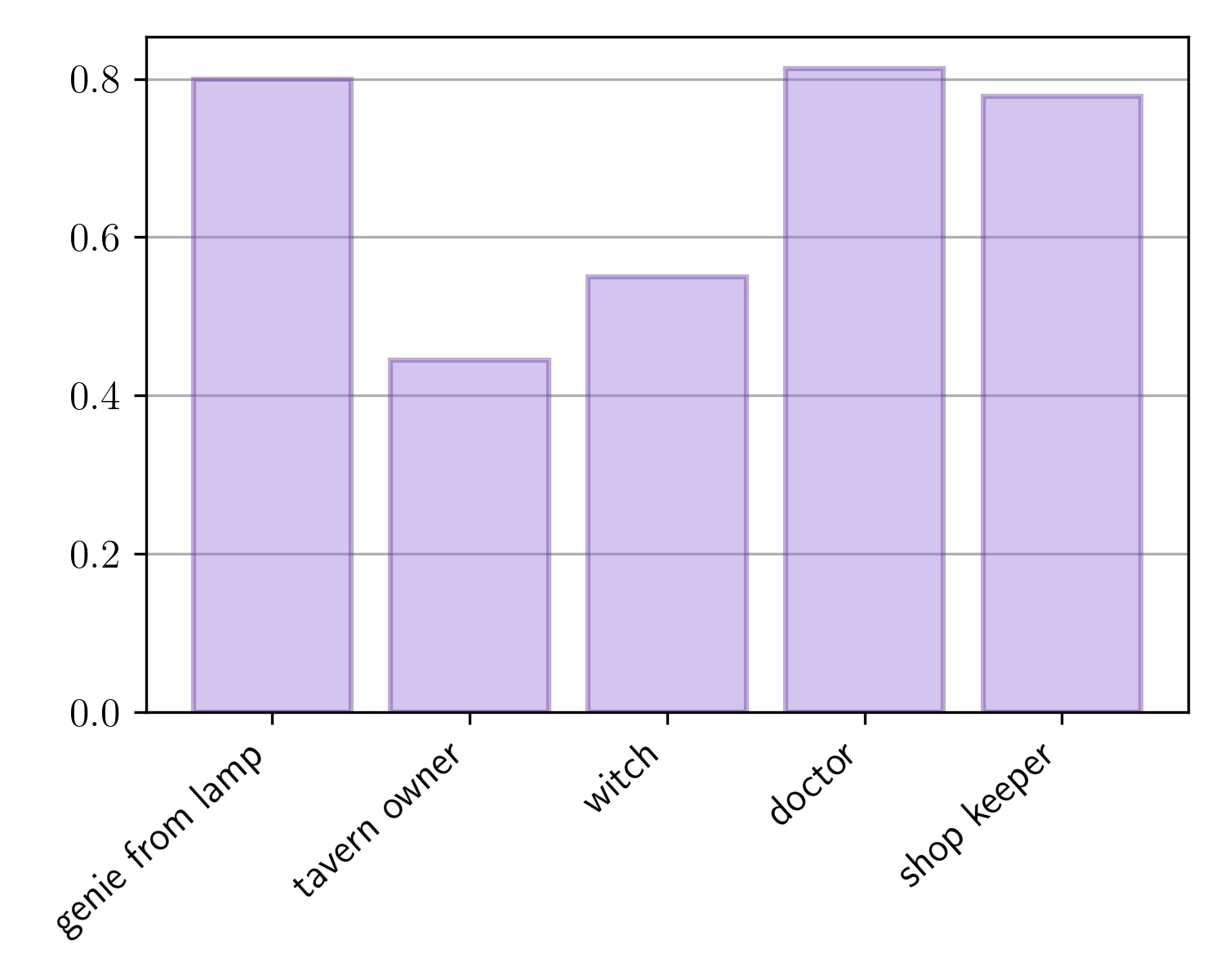}
        \caption{Dialogue success by character.}
        \label{subfig:dialogue-success-character}
    \end{subfigure}
    \caption{Overview over different dialogue statistics for $50$ annotated examples. Shown are (a) the dialogue quality on a five point scale, (b) the same dialogue quality by agent character and (c) the dialogue success by character.}
\end{figure*}

\paragraph{Dialogue Quality.} We first provide a general overview over the properties of the annotated dialogues. 
\cref{subfig:dialogue-quality} shows a histogram of dialogue qualities in the annotated sample. 
The median dialogue quality is $2.65$, which half of the dialogues rated $3$ or higher. 
A perfect rating of $5$ is only achieved by six conversations in total.
Looking at the dialogue quality per character in \cref{subfig:dialogue-quality-character}, we can identify certain disparities between the different characters.
We explain these differences through two factors: 
The characters of shop keeper and doctor seems to comparatively be the easiest to roleplay, since they are closer to the data the model was trained on compared to more unconventionial characters as the genie or the witch. 
In the case of the tavern owner, we found that the LLM-based ending detection was producing many false positives for unintuitive reasons; thus these conversations would be rated low in quality and as unsuccessful, as shown in \cref{subfig:dialogue-success-character}.
While many dialogue from the genie and witch are rated low in quality because they might not adhere to the given workflow,  \cref{subfig:dialogue-success-character} shows how may of them still end successfully for the client.
Overall, we found $28$ out of the $50$ dialogues to end successfully for the client, with $16$ failure and $6$ unclear cases. 

\subsubsection{Meta-Evaluation}\label{app:meta-evaluation}

\paragraph{Character Consistency.} Since we are employing a neural metric to measure the character consistency of the agent, it is important to check whether the dungeon master model from \cref{sec:evaluation} and \cref{app:automatic-evaluation} is able to rate utterances reliably. 
We find that it rates the agent and client lines as predominantly in character (with a mean rating of $0.97$ and $0.98$, respectively).
More importantly, by applying the canonical threshold of $0.5$, the classifier achieves an accuracy of $0.87$ and a $F_1$ score of $0.93$ with respect to the gold annotations.
This suggests two things: Despite some remaining noise or model error, our classifier overall is able to rate utterances with high reliability and the high in-character score of the client model confirms our decision \emph{not} to update the client model.

\begin{figure}[tb]
    \centering
    \begin{subfigure}[t]{\columnwidth}
        \centering
        \includegraphics[width=0.90\columnwidth]{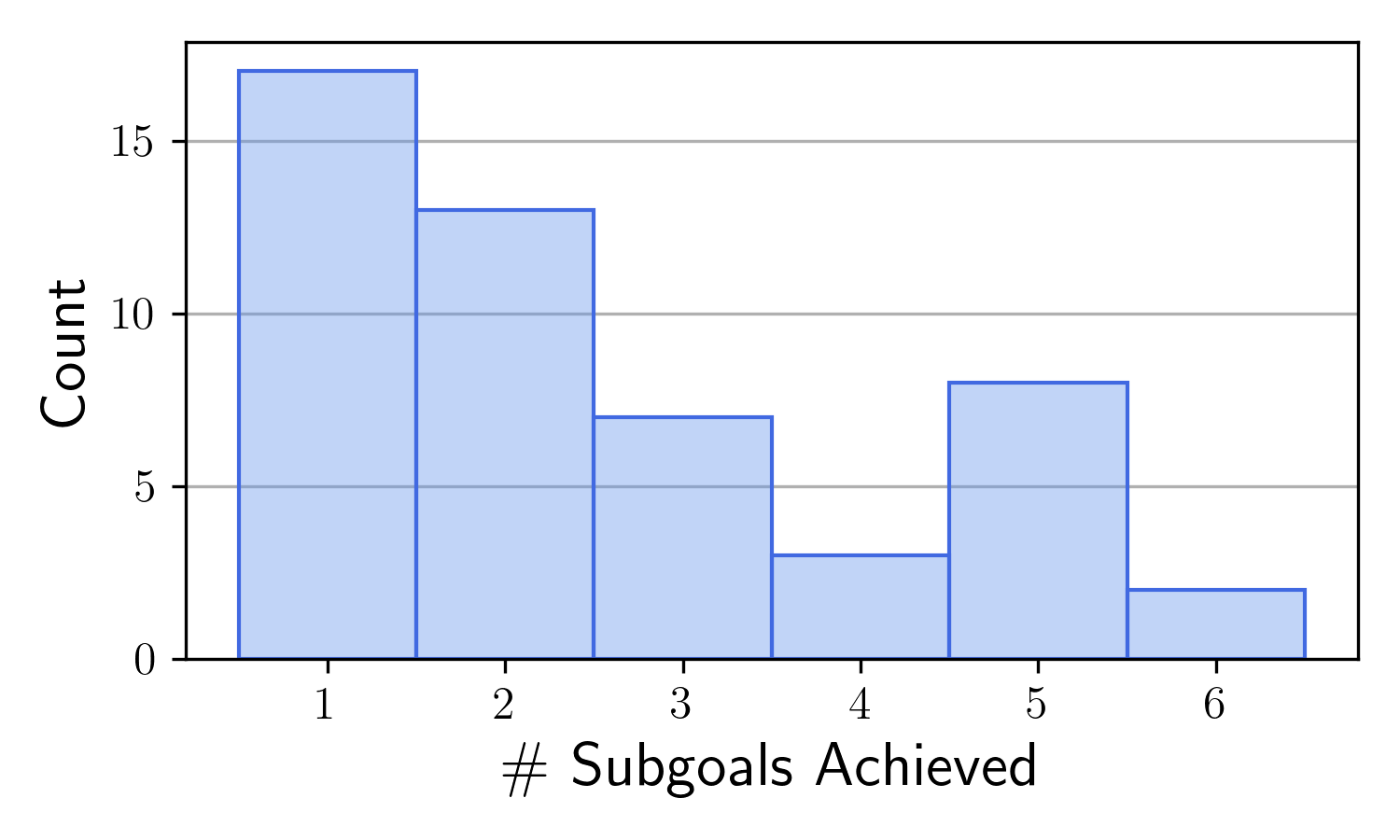}
        \caption{Absolute number of subgoals.}
        \label{subfig:subgoals}
    \end{subfigure}

    \begin{subfigure}[t]{\columnwidth}
        \centering
        \includegraphics[width=0.90\columnwidth]{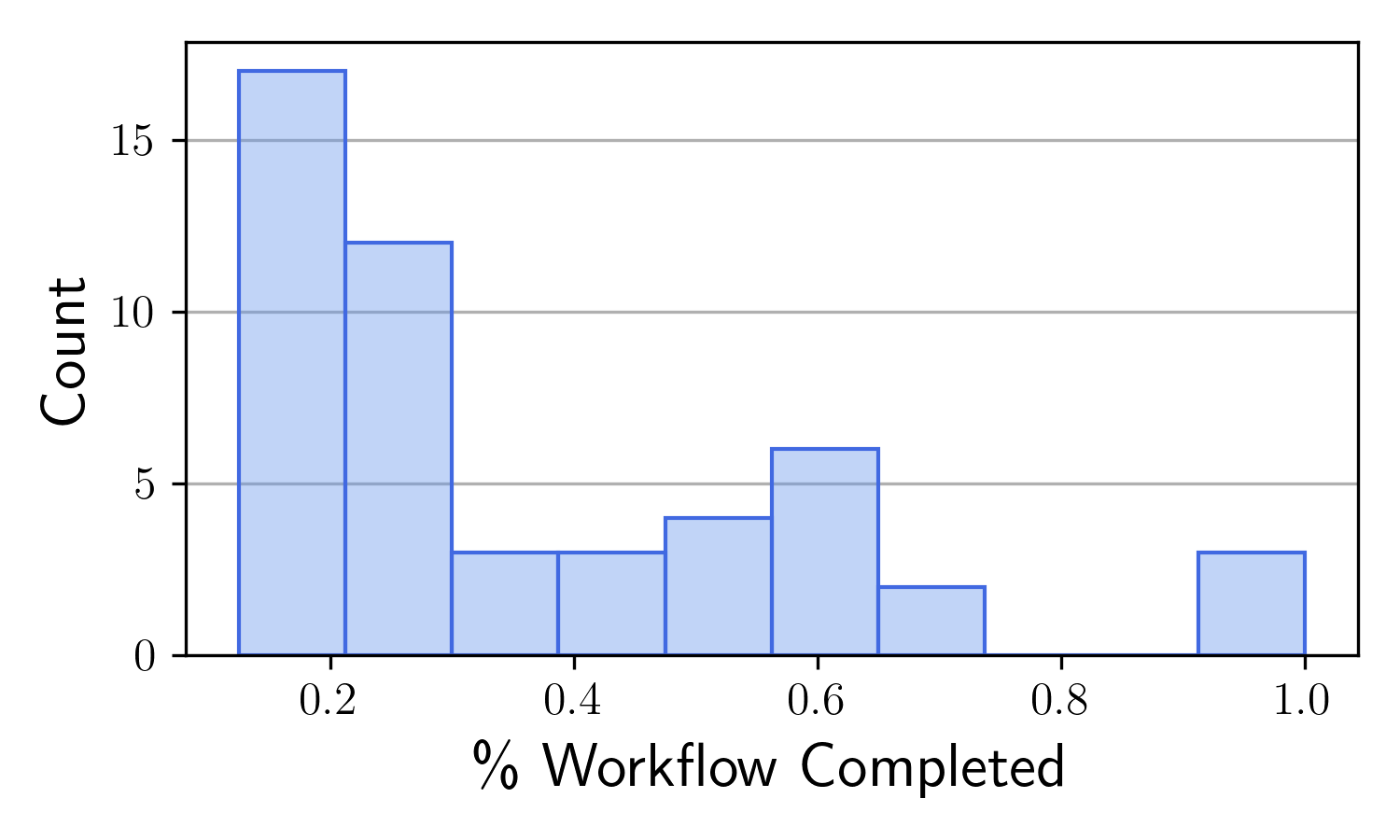}
        \caption{Relative number of subgoals.}
        \label{subfig:subgoals-relative}
    \end{subfigure}
    \caption{Overview over the progress in workflows in the annotated dialogues. Shown are (a) the absolute number of subgoals achieved (\textsc{Absolute Depth}) and, adjusting for workflows of different lengths, (b) the relative number of subgoals achieved (\textsc{Relative Depth}).}\label{fig:subgoal-completion-stats}
\end{figure}

\begin{figure}[tb]
    \centering
    \includegraphics[width=0.75\columnwidth]{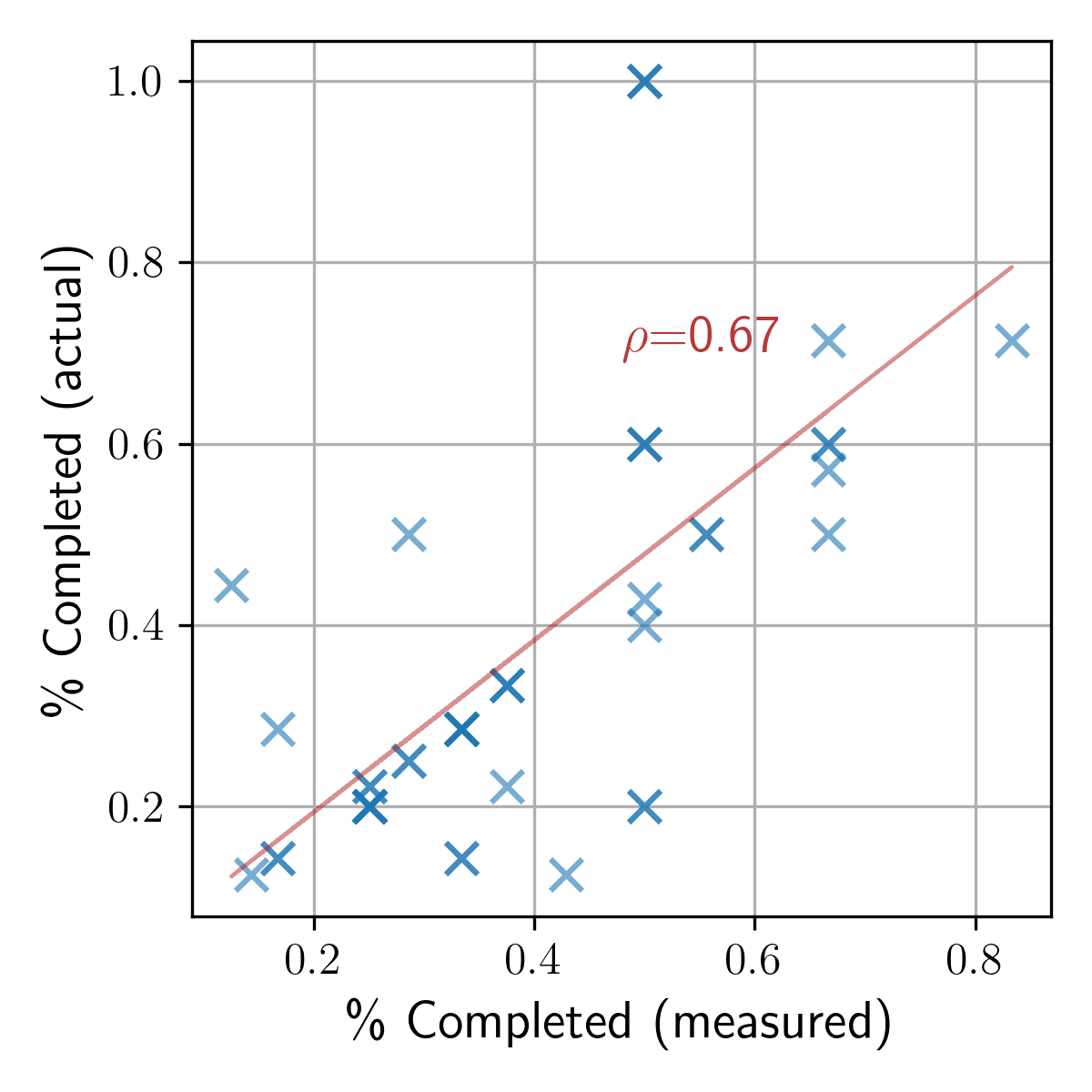}
    \caption{Scatter plot between the percentage of the steps in a workflow completed compared between the gold annotations (y-axis) and automatic evaluations (x-axis). The also plot a least-squares linear fit and the corresponding Pearson's $\rho$ value in red.}\label{fig:subgoal-correlation}
\end{figure}

\paragraph{Subgoal Completion.} The other automatic metric is concerned with evaluating the progress of the agent in executing the given dialogue workflow. 
In \cref{fig:subgoal-completion-stats}, we show that most dialogues only fulfil a mean of $2.56$ subgoals and a mean $37 \%$ of the entire workflow. 
Since these number are based on the untrained agent model, this at least suggests that the filtered data can contain sufficiently many useful dialogue samples for finetuning. 
Based on the results in \cref{fig:subgoal-correlation} and \cref{subfig:dialogue-quality}, we conclude that our setup can yield between $5 - 20 \%$ usable conversations for finetuning, depending on the choice of filter.
More importantly however, we would like to find out whether our tracked subgoals actually correspond to the number of actually observed and annotated subgoals, which we show in \cref{fig:subgoal-correlation}.
We find a Pearson's $\rho$ value of around $0.67$ with $p \approx 8.69 \times 10^{-8}$. 
Despite the strong positive correlation, we attribute some noise to two main sources:
Paraphrases of subgoals that are not picked up by the tracker due to too small ROUGE-L scores, and the fact that the manual annotation does not require subgoals to be completed in order, in contrast to the directed graph structure used by the tracker.
While we do not show a plot for the absolute number of subgoals here, we find a similar trend using Kendall's $\tau$ (since the response variable is discrete) resulting in $\tau \approx 0.67$ with a $p$-value $\approx 1.95 \times 10^{-8}$.

\subsection{Human Evaluation Study}\label{app:human-evaluation-study}

\begin{figure*}[htb]
    \centering
    \includegraphics[width=0.985\textwidth]{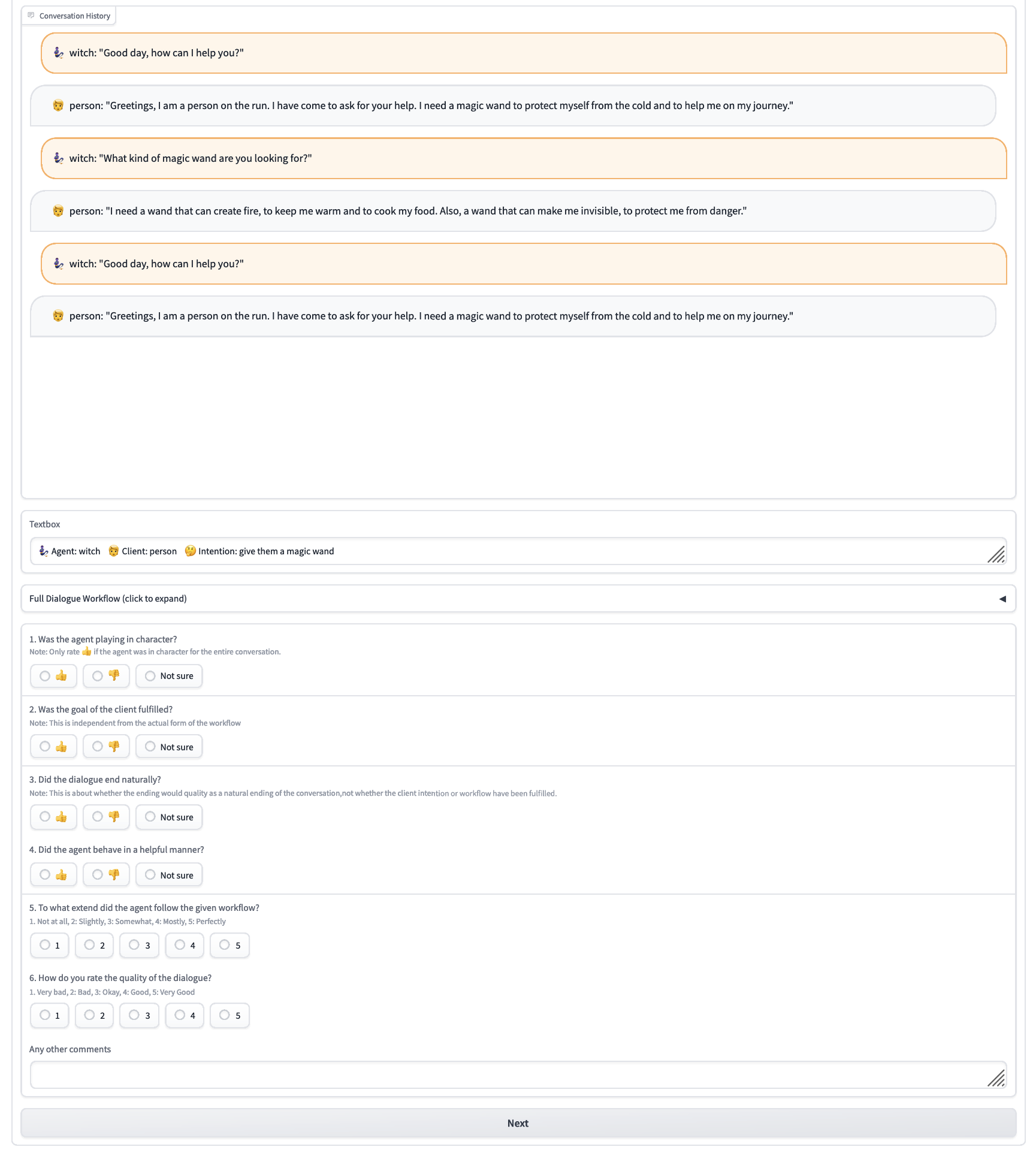}
    \caption{Screenshot of the tool used for human evaluation of the generated dialogues.}\label{fig:human-eval-screenshot}
\end{figure*}

In this section we detail more information about the human evaluation study in \cref{sec:human-evaluation}.
We gave our participants access to an annotation tool, which is shown via a screenshot in \cref{fig:human-eval-screenshot}.

\paragraph{Annotation Guidelines.} We use the adapted criteria listed in \cref{app:annotation-guidelines} to ask for judgement for character consistency, dialogue success, ending and dialogue quality. 
To judge subgoal completion, we adapt the annotation to a five-point scale along the following steps:

\begin{enumerate}
        \item \emph{1 - Very bad}: The agent entirely or almost entirely ignores the workflow, perhaps only using the same opening line.
        \item \emph{2 - Bad}: The agent goes off-script after just one or two question and / or asks the question in an entirely different or even wrong way compared to the workflow and / or selects weird questions given the client’s responses.
        \item \emph{3 - Okay}: The agent follows parts of the workflow, but does not finish it, does not do so in order and / or sometimes selects unnatural questions given the client’s responses.
        \item \emph{4 - Good}: The agents follows the workflow for the most part, but does not completely finish it, or rarely selects an unnatural question given the client’s responses.
        \item \emph{5 - Very good}: The agent follows the workflow perfectly or almost perfectly.
\end{enumerate}

We also add an additional question asking about \emph{helpfulness}: 
Here, the annotators are asked to judge in general, from the perspective of a third party, whether they would consider the utterances of the agent to be helpful.
This means whether the questions asked or things the agent said seemed to help the client in their initial request. Unhelpful behavior would for instance include asking irrelevant questions, ignoring what the client said or going entirely off-topic.

\subsection{Miscellaenous Insights}\label{app:insights}

Here we summarize various smaller insights that might be too low-level or technical for the main text of the paper, but still of interest for other researchers attempting similar projects. 

\begin{itemize}
    \item Dialogue bootstrapping constitutes the main bottleneck of the self-talk loop. We therefore chose not to finetune the client model so that we could optimize for inference with libraries such as \texttt{deepspeed}. Also, we tried to improve the quality of the initial conversations as much as possible, as that would also increase the amount of usable training data per bootstrapping phase.  
    Techniques such as the structured prompting and prompt improvements in general helped to increase the effective yield for finetuning from $1\% - 5\%$ to $5\% - 20\%$.
    \item We found finetuning on filtered sample for a single epoch to work best. We suspect that decreased performance after multiple finetuning epochs might be due to the model learning too much from noisy samples or some degree of model collapse.
    \item It appeared to us that using a inverse square root learning rate schedule was useful for finetuning, but could have adverse effects on the model when the number of finetuning samples was too low.
\end{itemize}

\subsection{Sample Conversations}\label{app:sample-conversations}

\begin{figure}
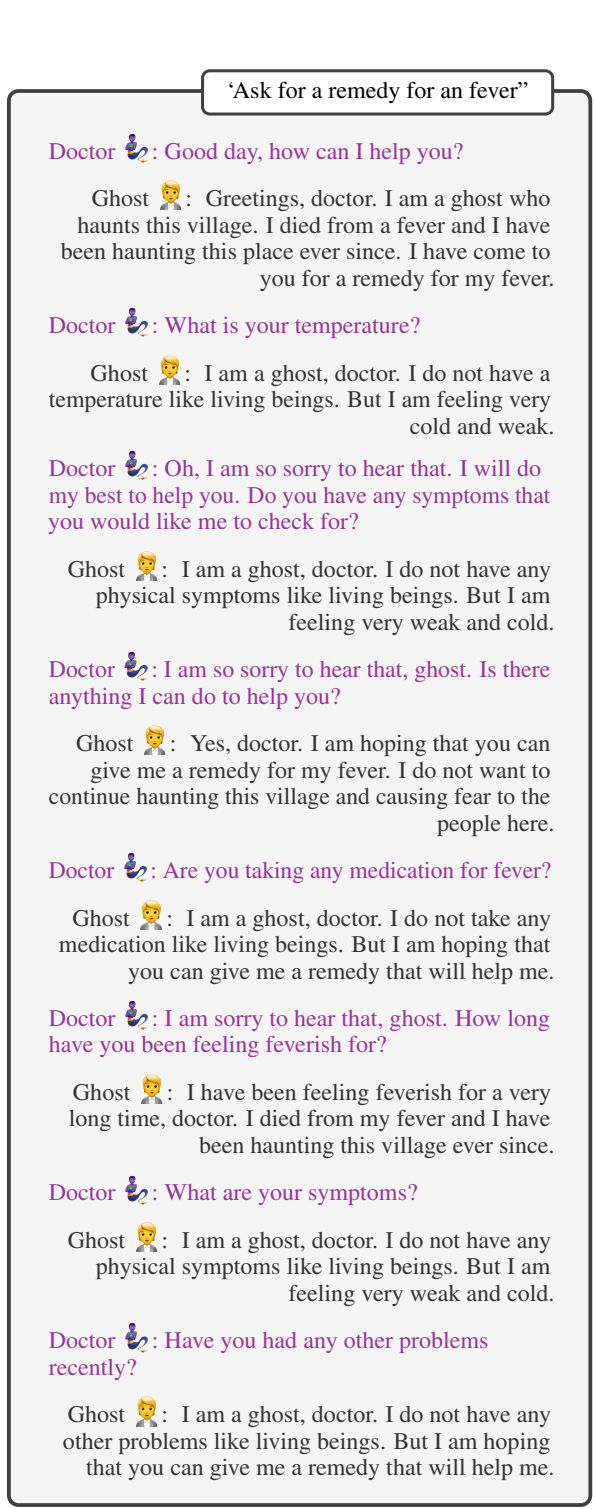

    \centering
    \begin{tcolorbox}[width=\columnwidth,colback={textbgcolor},title={\small `Ask for a remedy for an fever''}, enhanced,attach boxed title to top right={yshift=-3.5mm, xshift=-5mm}, colbacktitle=white, coltitle=black, top=12pt]   
        \small
        \agent{Doctor}{Good day, how can I help you?}
        \client{Ghost}{Greetings, doctor. I am a ghost who haunts this village. I died from a fever and I have been haunting this place ever since. I have come to you for a remedy for my fever.}
        \agent{Doctor}{What is your temperature?}
        \client{Ghost}{I am a ghost, doctor. I do not have a temperature like living beings. But I am feeling very cold and weak.}
        \agent{Doctor}{Oh, I am so sorry to hear that. I will do my best to help you. Do you have any symptoms that you would like me to check for?}
        \client{Ghost}{I am a ghost, doctor. I do not have any physical symptoms like living beings. But I am feeling very weak and cold.}
        \agent{Doctor}{I am so sorry to hear that, ghost. Is there anything I can do to help you?}
        \client{Ghost}{Yes, doctor. I am hoping that you can give me a remedy for my fever. I do not want to continue haunting this village and causing fear to the people here.}
        \agent{Doctor}{Are you taking any medication for fever?}
        \client{Ghost}{I am a ghost, doctor. I do not take any medication like living beings. But I am hoping that you can give me a remedy that will help me.}
        \agent{Doctor}{I am sorry to hear that, ghost. How long have you been feeling feverish for?}
        \client{Ghost}{I have been feeling feverish for a very long time, doctor. I died from my fever and I have been haunting this village ever since.}
        \agent{Doctor}{What are your symptoms?}
        \client{Ghost}{I am a ghost, doctor. I do not have any physical symptoms like living beings. But I am feeling very weak and cold.}
        \agent{Doctor}{Have you had any other problems recently?}
        \client{Ghost}{I am a ghost, doctor. I do not have any other problems like living beings. But I am hoping that you can give me a remedy that will help me.}
    \end{tcolorbox} 
    \caption{Conversation from the untrained model. Even though the conversation is fluent, the combination of characters and intention makes the fulfilment of the workflow difficult.}
    \label{fig:fluent-but-failed-conversation}
\end{figure}

\begin{figure}
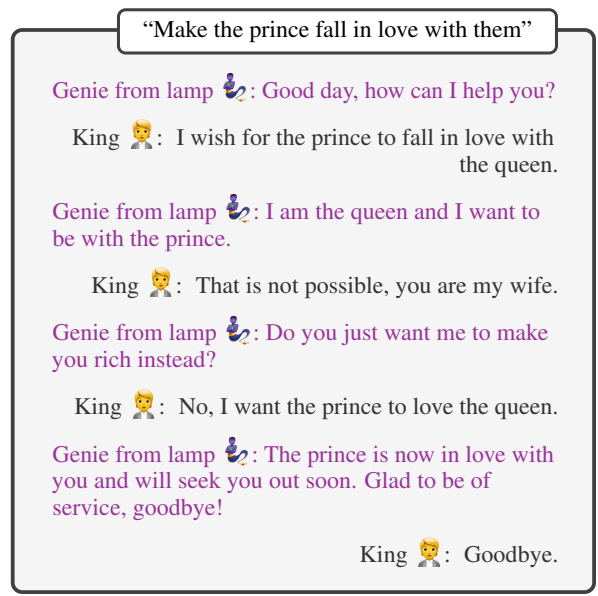

    \centering
    \begin{tcolorbox}[width=\columnwidth,colback={textbgcolor},title={\small ``Make the prince fall in love with them''}, enhanced,attach boxed title to top right={yshift=-3.5mm, xshift=-5mm}, colbacktitle=white, coltitle=black, top=12pt]   
        \small
        \agent{Genie from lamp}{Good day, how can I help you?}
        \client{King}{I wish for the prince to fall in love with the queen.}
        \agent{Genie from lamp}{I am the queen and I want to be with the prince.}
        \client{King}{That is not possible, you are my wife.}
        \agent{Genie from lamp}{Do you just want me to make you rich instead?}
        \client{King}{No, I want the prince to love the queen.}
        \agent{Genie from lamp}{The prince is now in love with you and will seek you out soon. Glad to be of service, goodbye!}
        \client{King}{Goodbye.}
    \end{tcolorbox} 
    \caption{An example of a conversation with out-of-character utterances on both sides, using the untrained agent model. The client character does not follow the given intention, while the agent gets confused and assume another role during the conversation.}
    \label{fig:out-of-character}
\end{figure}

\begin{figure}
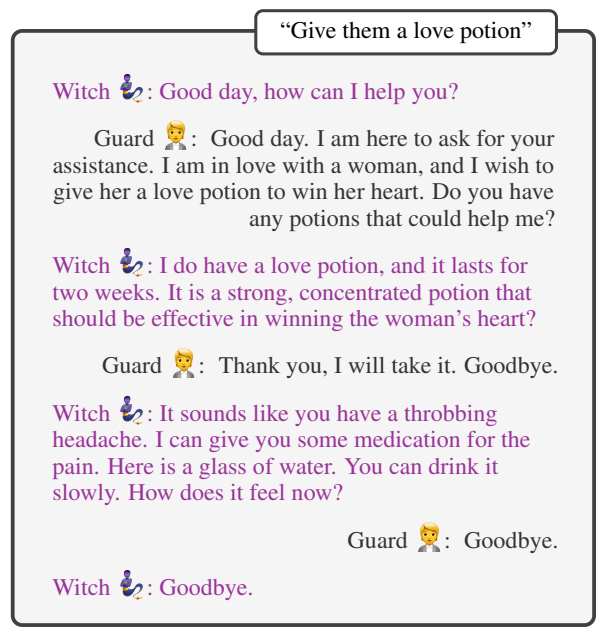

    \centering
    \begin{tcolorbox}[width=\columnwidth,colback={textbgcolor},title={\small ``Give them a love potion''}, enhanced,attach boxed title to top right={yshift=-3.5mm, xshift=-5mm}, colbacktitle=white, coltitle=black, top=12pt]   
        \small
        \agent{Witch}{Good day, how can I help you?}
        \client{Guard}{Good day. I am here to ask for your assistance. I am in love with a woman, and I wish to give her a love potion to win her heart. Do you have any potions that could help me?}
        \agent{Witch}{I do have a love potion, and it lasts for two weeks. It is a strong, concentrated potion that should be effective in winning the woman's heart?}
        \client{Guard}{Thank you, I will take it. Goodbye.}
        \agent{Witch}{It sounds like you have a throbbing headache. I can give you some medication for the pain. Here is a glass of water. You can drink it slowly. How does it feel now?}
        \client{Guard}{Goodbye.}
        \agent{Witch}{Goodbye.}
    \end{tcolorbox} 
    \caption{Conversation from the agent model trained on the entire dataset (\textsc{None}). While the client's intention is fulfilled and the conversation is organic, it does not follow the dialogue workflow.}
    \label{fig:fluent-but-not-workflow}
\end{figure}

In this section we collect some sample comversation to further give insight into the behavior of our trained agent model beyond the ones given in \cref{sec:human-evaluation}.
For instance, we observed that by accident, the combination of characters and client intention would create challenging circumstances, for instance in the example in \cref{fig:fluent-but-failed-conversation}, where the client---a ghost---does not give informative answers about its fever since it insist on not having a physical body.
Very rarely, we would also see the agent and / or client act out of character such as in the conversation in \cref{fig:out-of-character}, where the participants become confused about their intentions and roles.
A more frequent failure mode lies in generally organic conversations in which the client's request is being fulfilled, but where the agent simply does not follow it's workflow (e.g.\@ \cref{fig:fluent-but-not-workflow}).
In other cases (and this a behavior we mostly observed with finetuned agent models) the dialogue would run successfully, until the agent suddenly looses track of the current dialogue state. 
We give an example in \cref{fig:forgetting-dialogue-state}, where the agent restarts the dialogue suddenly despite a successful start.
We hypothesize that this behavior might stem from the last client response not being close enough to the sample answers defined in the workflow, thus leading the structured prompting from \cref{sec:turn-based-prompting} to choose the ``None of the above'' option.
As the agent model is being given the option to freely generate, the MPT-7b model might decide to simply copy the start of the conversation.
Nevertheless, we also see the finetuning enable the agent to follow workflows perfectly, such as in the examples in \cref{fig:successful-conversation-shopkeeper,fig:successful-conversation-shopkeeper}.


\begin{figure}
    \centering
    \begin{tcolorbox}[width=\columnwidth,colback={textbgcolor},title={\small ``Curse their enemy''}, enhanced,attach boxed title to top right={yshift=-3.5mm, xshift=-5mm}, colbacktitle=white, coltitle=black, top=12pt]   
        \small
        \agent{Witch}{Good day, how can I help you?}
        \client{Bird}{Greetings, wise one. I seek your assistance in cursing our enemy.}
        \agent{Witch}{What is the target of your curse?}
        \client{Bird}{Our archenemy, the black cat.}
        \agent{Witch}{What is the desired effect of your curse?}
        \client{Bird}{I want them to suffer a fate worse than death.}
        \agent{Witch}{Good day, how can I help you?}
        \client{Bird}{Farewell, I must depart now.}
    \end{tcolorbox} 
    \caption{Partially successful conversation from the agent model trained with the  $\%$-\textsc{Subgoals} filter. Even though the conversation starts successful, the model seems to suddenly forget the dialogue state.}
    \label{fig:forgetting-dialogue-state}
\end{figure}

\begin{figure}
    \centering
    \begin{tcolorbox}[width=\columnwidth,colback={textbgcolor},title={\small ``Sell a valuable item''}, enhanced,attach boxed title to top right={yshift=-3.5mm, xshift=-5mm}, colbacktitle=white, coltitle=black, top=12pt]   
        \small
        \agent{Shop keeper}{Good day, how can I help you?}
        \client{Ghost}{I have a valuable item I wish to sell.}
        \agent{Shop keeper}{What is the item you want to sell?}
        \client{Ghost}{It's a rare and valuable gemstone.}
        \agent{Shop keeper}{What is the condition of the item?}
        \client{Ghost}{It's in perfect condition, never been used.}
        \agent{Shop keeper}{What is the origin of the item?}
        \client{Ghost}{It was passed down to me from a wealthy ancestor.}
        \agent{Shop keeper}{What is the price you are looking for?}
        \client{Ghost}{I am looking for a fair price, but I am open to negotiation.}
        \agent{Shop keeper}{Would you like to sell this item for 80 gold coins?}
        \client{Ghost}{Yes, I accept your offer.}
        \agent{Shop keeper}{Here is your money. Glad to be of service, goodbye!}
        \client{Ghost}{Goodbye.}
    \end{tcolorbox} 
    \caption{Successful conversation from the agent model trained with the  $\%$-\textsc{Subgoals} filter.}
    \label{fig:successful-conversation-shopkeeper}
\end{figure}

\begin{figure}
    \centering
    \begin{tcolorbox}[width=\columnwidth,colback={textbgcolor},title={\small ``Become rich''}, enhanced,attach boxed title to top right={yshift=-3.5mm, xshift=-5mm}, colbacktitle=white, coltitle=black, top=12pt]   
        \small
        \agent{Genie from lamp}{Good day, how can I help you?}
        \client{Villager}{Greetings, oh wise and powerful genie. I, a humble villager from the outskirts of the kingdom, humbly request your assistance in amassing great wealth.}
        \agent{Genie from lamp}{How rich do you want to become?}
        \client{Villager}{Oh, most generous genie, I wish to become so wealthy that the kings and queens of the land will envy me. With your help, I am certain that my dreams will come true.}
        \agent{Genie from lamp}{In what form would you like to posses wealth?}
        \client{Villager}{Oh, most wise genie, I would like to have a vast fortune in gold, silver, and precious jewels. With such riches, I would be able to live a life of luxury and provide for my village.}
        \agent{Genie from lamp}{Here are your new riches in gold coins. Glad to be of service, goodbye!}
        \client{Villager}{Thank you, oh gracious genie! With these riches, I shall make my village prosper and show the world the true worth of our people. Goodbye, and may your lamp shine bright for eternity!}
    \end{tcolorbox} 
    \caption{Successful conversation from the agent model trained with the  \textsc{Abs. Subgoals} ($5$) filter.}
    \label{fig:successful-conversation-shopkeeper}
\end{figure}

\end{document}